\documentclass[10pt,twocolumn,letterpaper]{article}

\usepackage[pagenumbers]{cvpr} % To force page numbers, e.g. for an arXiv version

\usepackage{graphicx}
\usepackage{amsmath}
\usepackage{amssymb}
\usepackage{booktabs}
\usepackage{algorithm}
\usepackage{algpseudocode}
\usepackage{array}
\usepackage{xcolor}
\usepackage{adjustbox}
\usepackage{wrapfig}
\usepackage[table]{colortbl}
\usepackage{multirow}

\newcommand{\Lunivl}{\mathcal{L}_{\text{UniVL}}}
\newcommand{\Lalign}{\mathcal{L}_{\text{Align}}}
\newcommand{\Ljoint}{\mathcal{L}_{\text{Joint}}}
\newcommand{\Ldecoder}{\mathcal{L}_{\text{Decoder}}}
\newcommand{\Lmdecoder}{\mathcal{L}_{\text{M-Decoder}}}
\newcommand{\Lmalign}{\mathcal{L}_{\text{M-Align}}}
\newcommand{\Lcmlm}{\mathcal{L}_{\text{CMLM}}}
\newcommand{\Lcmfm}{\mathcal{L}_{\text{CMFM}}}
\newcommand{\Lmjoint}{\mathcal{L}_{\text{M-Joint}}}
\newcommand{\pri}{\mathcal{L}^\text{pri}}
\newcommand{\aux}{\mathcal{L}^\text{aux}}
\newcommand{\Laux}{\mathcal{L}^\text{aux}}
\newcommand{\bbR}{\mathbb{R}}
\newcommand{\Ell}{\boldsymbol{\ell}}
\newcommand{\LcVc}{\mathcal{L}_{\mathcal{V}}}
\newcommand{\Lct}{\mathcal{L}_{t}}
\newcommand{\Lreg}{\mathcal{L}^\text{reg}}
\DeclareMathOperator*{\argmin}{arg\,min}
\usepackage[pagebackref,breaklinks,colorlinks]{hyperref}

\usepackage[capitalize]{cleveref}
\crefname{section}{Sec.}{Secs.}
\Crefname{section}{Section}{Sections}
\Crefname{table}{Table}{Tables}
\crefname{table}{Tab.}{Tabs.}

\def\cvprPaperID{6324} % *** Enter the CVPR Paper ID here
\def\confName{CVPR}
\def\confYear{2023}

\begin{document}

%%%%%%%%% TITLE - PLEASE UPDATE
\title{MELTR: Meta Loss Transformer for\\ Learning to Fine-tune Video Foundation Models}

\newcommand*\samethanks[1][\value{footnote}]{\footnotemark[#1]}
\author{
Dohwan Ko\textsuperscript{\rm 1}\thanks{Equal contribution.}\hspace{0.4cm}
Joonmyung Choi\textsuperscript{\rm 1}\samethanks\hspace{0.4cm}
Hyeong Kyu Choi\textsuperscript{\rm 1}\hspace{0.4cm} \\
Kyoung-Woon On\textsuperscript{\rm 2}\hspace{0.4cm} 
Byungseok Roh\textsuperscript{\rm 2}\hspace{0.4cm}
Hyunwoo J. Kim\textsuperscript{\rm 1}\thanks{Corresponding author.}\vspace{0.3cm} \\
\textsuperscript{\rm 1}Department of Computer Science and Engineering, Korea University\hspace{0.4cm}\vspace{0cm} \textsuperscript{\rm 2}Kakao Brain \\
\tt\small \{ikodoh, pizard, imhgchoi, hyunwoojkim\}@korea.ac.kr \vspace{0cm}
\\\tt\small \{kloud.ohn, peter.roh\}@kakaobrain.com \vspace{0cm}
}

\maketitle

%%%%%%%%% ABSTRACT
\begin{abstract}
    Foundation models have shown outstanding performance and generalization capabilities across domains.
    Since most studies on foundation models mainly focus on the pretraining phase, a naive strategy to minimize a single task-specific loss is adopted for fine-tuning.
    However, such fine-tuning methods do not fully leverage other losses that are potentially beneficial for the target task.
    Therefore, we propose \textbf{ME}ta \textbf{L}oss \textbf{TR}ansformer (\textbf{MELTR}), a plug-in module that automatically and non-linearly combines various loss functions to aid learning the target task via auxiliary learning.
    We formulate the auxiliary learning as a bi-level optimization problem and present an efficient optimization algorithm based on Approximate Implicit Differentiation (AID).
    For evaluation, we apply our framework to various video foundation models (UniVL, Violet and All-in-one), and show significant performance gain on all four downstream tasks: text-to-video retrieval, video question answering, video captioning, and multi-modal sentiment analysis.
    Our qualitative analyses demonstrate that MELTR adequately `transforms' individual loss functions and `melts' them into an effective unified loss.
    Code is available at \url{https://github.com/mlvlab/MELTR}.
\end{abstract}
%%%%%%%%% BODY TEXT
\section{Introduction}
Large-scale models trained on a huge amount of data have gained attention due to their adaptability to a wide range of downstream tasks. 
As introduced in~\cite{bommasani2021opportunities}, deep learning models with the generalizability are referred to as foundation models. 
In recent years, several foundation models for various domains have been proposed (\textit{e.g.}, \cite{bertasius2021space,brown2020language} for natural language processing, \cite{radford2021learning,jia2021scaling} for images and language, and \cite{akbari2021vatt,luo2020univl,zellers2021merlot} for videos) and they mainly focus on \textit{pretrain} the model often with various \textit{multiple} pretext tasks. 
On the other hand, strategies for fine-tuning on downstream tasks are less explored.
For instance, a recently proposed video foundation model UniVL~\cite{luo2020univl} is pretrained with a \textit{linear} combination of several pretext tasks such as text-video alignment, masked language/frame modeling, and caption generation.
However, like other domains, fine-tuning is simply performed by minimizing a \textit{single} target loss. 
Other potentially beneficial pretext tasks have remained largely unexplored for fine-tuning. 

Auxiliary learning is a natural way to utilize multiple pretext task losses for learning.
Contrary to multi-task learning that aims for generalization across tasks, auxiliary learning focuses only on the primary task by taking advantage of several auxiliary tasks.
Most auxiliary learning frameworks~\cite{liebel2018auxiliary,Toshniwal2017} manually selected auxiliary tasks, which require domain knowledge and may not always be beneficial for the primary task.
To automate task selection, meta learning was integrated into auxiliary learning~\cite{liu2019self,navon2020auxiliary,shu2019meta}.
Here, the model learns to adaptively leverage multiple auxiliary tasks to assist learning of the primary task.
Likewise, the pretext task losses can be unified into a single auxiliary loss to be optimized in a way that helps the target downstream task.

To this end, we propose Meta Loss Transformer (MELTR), a plug-in module that \textit{automatically} and \textit{non-linearly} transforms various auxiliary losses into a unified loss.
MELTR built on Transformers~\cite{vaswani2017attention} takes the target task loss as well as pretext task losses as input and  learns their relationship via self-attention.
In other words, MELTR learns to fine-tune a foundation model by combining the primary task with multiple auxiliary tasks, and this can be viewed as a meta-learning (or `learning-to-learn') problem.
Similar to meta-learning-based auxiliary learning frameworks~\cite{hwang2020self,shu2019meta}, this can be formulated as a bi-level optimization problem, which generally involves a heavy computational cost due to the second-order derivative and its inverse, \textit{e.g.}, the inverse Hessian matrix. 
To circumvent this, we present an efficient training scheme that approximates the inverse Hessian matrix.
We further provide empirical analyses on the time-performance trade-off of various optimization algorithms.
 
To verify the generality of our proposed method, we apply it to three video foundation models: UniVL~\cite{luo2020univl}, Violet~\cite{fu2021violet}, and All-in-one~\cite{wang2022all}.
These foundation models are originally pretrained with a linear combination of several pretext tasks such as text-video alignment, masked language/frame modeling, and caption generation.  
We experiment by fine-tuning on the text-to-video retrieval, video question answering, video captioning, and multi-modal sentiment analysis task with five datasets: YouCook2, MSRVTT, TGIF, MSVD, and CMU-MOSI.
For each task and dataset, our MELTR improves both previous foundation models and task-specific models by large margins.
Furthermore, our extensive qualitative analyses and ablation studies demonstrate that MELTR effectively learns to non-linearly combine pretext task losses, and adaptively re-weights them for the target downstream task.

\noindent To sum up, our \textbf{contributions} are threefold:
\begin{itemize}
    \item[\textbullet] We propose \textbf{ME}ta \textbf{Lo}ss \textbf{TR}ansformer (\textbf{MELTR}), a novel fine-tuning framework for video foundation models. 
    We also present an efficient optimization algorithm to alleviate the heavy computational cost of bi-level optimization.
    \item[\textbullet] We apply our framework to three video foundation models in four downstream tasks on five benchmark video datasets, where MELTR significantly outperforms the baselines fine-tuned with single-task and multi-task learning schemes.
    \item[\textbullet] We provide in-depth qualitative analyses on how MELTR non-linearly transforms individual loss functions and combines them into an effective unified loss for the target downstream task.
\end{itemize}
\section{Related work}
\noindent \textbf{Video foundation models.}
With sufficient computational power and an abundant source of data, there have been attempts to build a single large-scale foundation model that can be adapted to diverse downstream tasks.
Along with the success of foundations models in the natural language processing domain~\cite{brown2020language,chen2021evaluating,devlin2019bert} and in computer vision~\cite{bertasius2021space,jia2021scaling,radford2021learning}, video data has become another data type of interest, as it has grown in scale due to numerous internet video-sharing platforms.
Accordingly, several methods to train a video foundation model have been proposed.
Due to the innate multi-modality of video data, \textit{i.e.}, a combination of visual $\cdot$ vocal $\cdot$ textual context, most works have centered around the variations of the cross-modal attention mechanism \cite{akbari2021vatt,bertasius2021space,gabeur2020multi,luo2020univl,neimark2021video,tan2021look,wei2020multi,yang2021taco}.
In addition, as most video data lack proper labels or descriptions, contrastive learning methods were studied to learn meaningful feature representations or enhance video-text alignment in a self-supervised manner \cite{akbari2021vatt,kuang2021video,luo2020univl,yang2021taco}.

More specifically, MERLOT \cite{zellers2021merlot} proposed a multi-modal representation learning method for visual commonsense reasoning, which also performed well in twelve video reasoning tasks.
VATT \cite{akbari2021vatt} introduced a multi-modal learning method via contrastive learning. 
The pre-trained model performed well in a variety of vision tasks from image classification to video action recognition and zero-shot video retrieval.
Another representative work, UniVL \cite{luo2020univl} proposed a straightforward pre-training method with auxiliary loss functions. 
After fine-tuning on a specific task, the pre-trained model performed outstandingly in a wide range of tasks of text-to-video retrieval, action segmentation, action step localization, video sentiment analysis, and video captioning.
Other foundation models for multiple video tasks include \cite{li2020hero,sun2019learning,sun2019videobert,zhu2020actbert,fu2021violet,wang2022all}. 

\noindent \textbf{Auxiliary learning.}
In order to enhance the performance of one or a multitude of primary tasks, auxiliary learning methods can be incorporated.
\cite{ruder2017overview} introduced Multi-task learning (MTL) to the deep neural networks by training a single model with multiple task losses to assist learning on the main task.
Such a method is generally adapted to pre-train the foundation models in the self-supervised manner~\cite{li2020hero,sun2019learning,sun2019videobert,zhu2020actbert,fu2021violet,wang2022all}.
However, these various pretext task losses used in the pre-training phase are ignored in the fine-tuning phase, and only the primary task loss is minimized.

Recently, meta-learning methods have been introduced for auxiliary learning.
\cite{liu2019self,navon2020auxiliary,shu2019meta} proposed a meta-learning method in which the model learns auxiliary tasks to generalize well to unseen data. 
In these settings, a separate subset of data is held out as the primary task, while the others are used as auxiliary tasks that aid the primary task's performance.
Similar methods were adopted for computer vision tasks such as semantic segmentation \cite{xu2021leveraging}.
Other domain applications include navigation tasks with reinforcement learning \cite{ye2021auxiliary}, or self-supervised learning methods on graph data \cite{hwang2020self}.
\section{Preliminaries}
We briefly introduce UniVL~\cite{luo2020univl}, a video foundation model used as one of baselines for our learning method.
We also explain two types of optimization schemes for bi-level optimization problems which commonly occur in meta-learning and auxiliary learning.

\begin{figure}[t] 
    \centering
    \begin{subfigure}[t]{0.49\linewidth}
        \includegraphics[width=\linewidth]{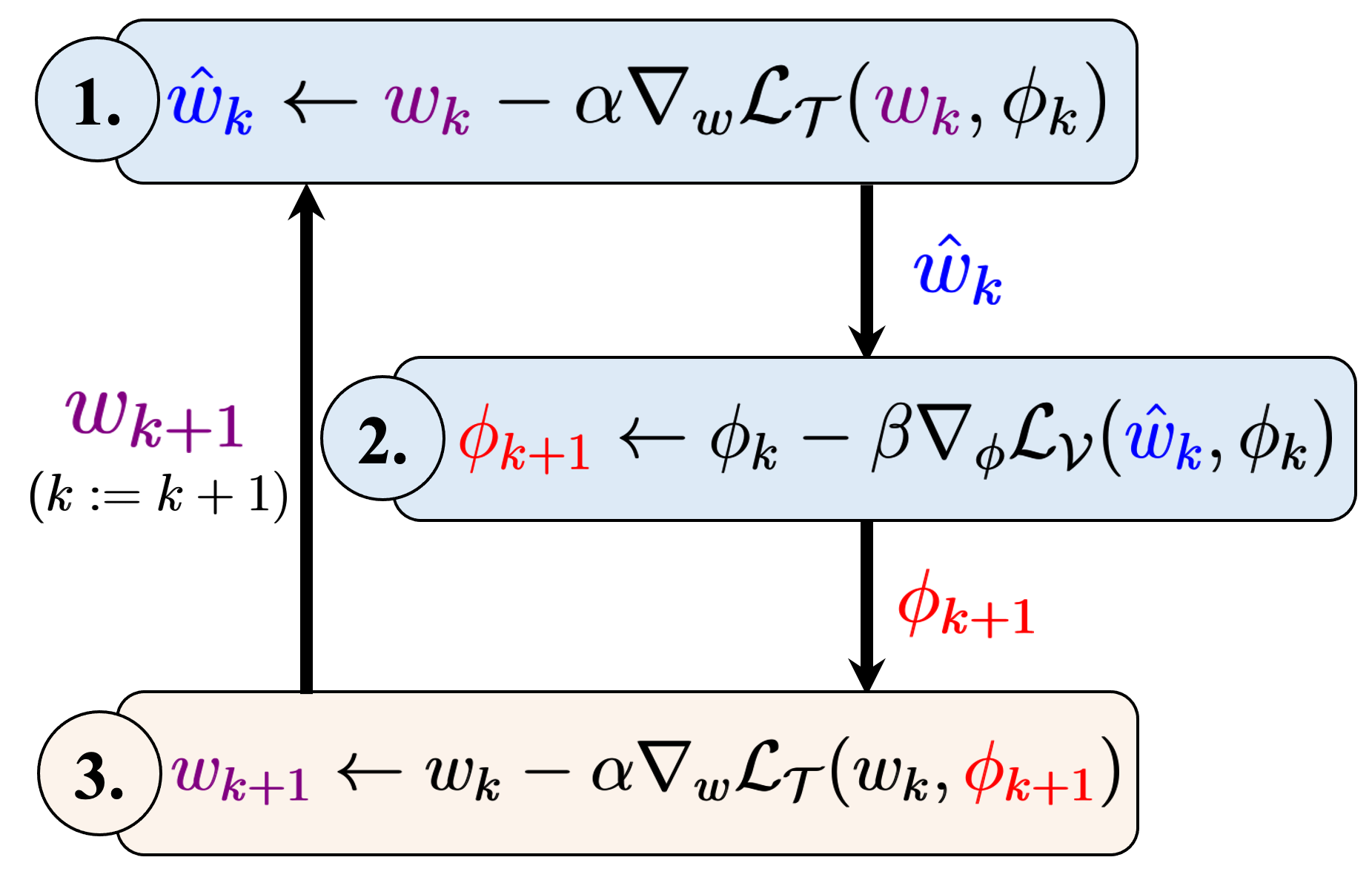}
        \caption{Iterative Differentiation}
        \label{fig:itd}
    \end{subfigure}
    \begin{subfigure}[t]{0.44\linewidth}
        \includegraphics[width=\linewidth]{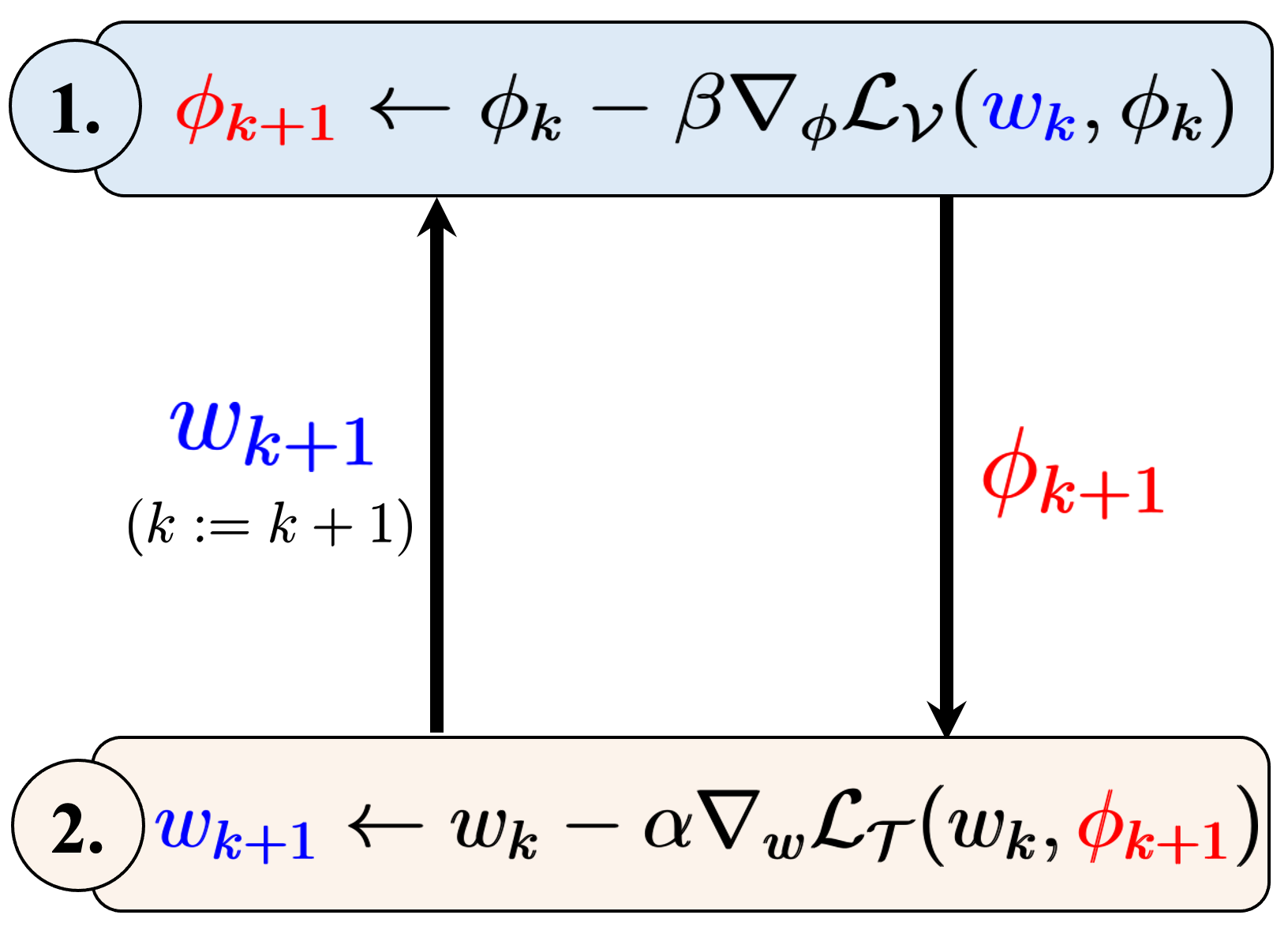}
        \caption{Approximate Implicit Diff.}
        \label{fig:aid}
    \end{subfigure}
    \caption{\textbf{Comparison of ITD and AID.} 
    The blue boxes indicate upper-level optimization, whereas the yellow boxes refer to lower-level optimization.
    (a) ITD defines a fixed-point parameter $\hat{w}_k$ for upper-level optimization. 
    To avoid visual clutter, iteration number for $\color{blue}\hat{w}_k$ update is set to 1. 
    (b) AID uses a 2-step optimization scheme which optimizes the upper-level decision vector in a single step via IFT. 
    Both algorithms can be computed efficiently with automatic differentiation~\cite{baydin2018automatic,derivatives2000principles}.
    }
    \label{fig:itd_aid}
\end{figure}
\subsection{UniVL}
\label{subsec:univl}
UniVL~\cite{luo2020univl} is a video foundation model pre-trained on the HowTo100M~\cite{miech2019howto100m} dataset via multi-modal self-supervised learning.
It is a unified video and language pre-training model for both video understanding and text generation tasks.
It consists of four transformer-based modules (two single-modal encoders, a cross-modal encoder, and a decoder).
It is pre-trained with five pretext tasks including the video-text joint ($\Ljoint$), the conditioned masked language model (CMLM; $\Lcmlm$), the conditioned masked frame model (CMFM; $\Lcmfm$), a video-text alignment ($\Lalign$), and the language generation task ($\Ldecoder$).
UniVL trains the model simultaneously for five pretext tasks by optimizing the sum of pretext loss functions given as:
\begin{align}
    \Lunivl &= \Ljoint + \Lcmlm + \Lcmfm + \Lalign + \Ldecoder.
\end{align}
Although UniVL minimizes \textit{multiple} pretext loss functions during pre-training, 
it optimizes only \textit{one} target task loss for fine-tuning, \eg, $\Lalign$ for video retrieval and $\Ldecoder$ for video captioning.
That is, other loss functions, which are potentially helpful for the target downstream task, are not utilized during fine-tuning.
This observation motivates our framework that automatically learns how to combine multiple losses for fine-tuning.
This can be viewed as hyperparameter optimization via meta-learning.
\subsection{Bi-level optimization and hypergradient approximation}
Bi-level optimization commonly arises in meta-learning and hyperparameter optimization.
One general class of bi-level problems is given as:
\begin{equation}
    \begin{split}
        & \min_{\phi \in \mathrm{\Phi}} f(\phi) := \mathcal{L}_\mathcal{V}(w^*(\phi), \phi) \\
        \text{s.t.} & \:\: w^*(\phi) = \argmin_{w} \mathcal{L}_\mathcal{T}(w, \phi),
    \end{split}
    \label{eq:bilevel1}
\end{equation}
where $\phi \in \mathrm{\Phi}$ is an upper-level decision vector, and $w \in \bbR^d$ is a lower-level decision vector. 
 $\mathcal{L}_\mathcal{V}:\bbR^d \times \mathrm{\phi} \rightarrow \bbR$ and $\mathcal{L}_\mathcal{T}: \bbR^d \times \mathrm{\phi} \rightarrow \bbR$ are upper-level and lower-level loss functions, respectively. 
For instance, in hyperparameter optimization, $\phi$ is a set of hyperparameters and $w$ is model parameters.
$\mathcal{L}_\mathcal{T}$ and $\mathcal{L}_\mathcal{V}$ can be mapped to training and validation loss functions, respectively.

Grazzi \etal. \cite{grazzi2020iteration} have investigated bi-level optimization algorithms from the {\em hypergradient} approximation perspective. 
Hypergradient is the gradient of upper-level objective (\textit{i.e.}, $\nabla \LcVc$) and it is used for updating upper-level decision vector $\phi$.
Popular approaches in the literature~\cite{grazzi2020iteration} can be categorized into two groups: Iterative Differentiation (ITD) and Approximate Implicit Differentiation (AID).

\noindent \textbf{Iterative Differentiation}~\cite{finn2017model,franceschi2017forward,franceschi2018bilevel,maclaurin2015gradient,shu2019meta}.
This algorithm unrolls the upper-level optimization into two stages by defining a fixed-point parameter, $\hat{w}_k(\phi)$, where $k$ denotes the upper-level optimization step.
$\hat{w}_k(\phi)$ is derived by taking iterative learning steps from $w_k$.
Then, assuming this as a contraction mapping with respect to $w_k$~\cite{grazzi2020iteration}, the hypergradient $\nabla \mathcal{L}_\mathcal{V}(w_k, \phi_k)$ can be approximated with $\nabla \mathcal{L}_\mathcal{V}(\hat{w}_k, \phi_k)$, as in Figure \ref{fig:itd_aid}(a) step 2.
Then, with the updated upper-level decision vector $\phi_{k+1}$, model parameter $w_{k+1}$ is computed in the final step 3. 

\noindent \textbf{Approximate Implicit Differentiation}~\cite{lorraine2020optimizing,pedregosa2016hyperparameter,rajeswaran2019meta}.
In this optimization scheme, the hypergradient $\nabla \mathcal{L}_\mathcal{V}(w_k, \phi_k)$ is factorized by the implicit function theorem (IFT). 
This is then solved with a 2-step algorithm which requires inverse Hessian computation (Figure \ref{fig:itd_aid}(b)).
In practice, the inverse Hessian matrix is generally approximated to avoid its computation overhead of $O(n^3)$.
Then, similarly to ITD, model parameter $w_{k}$ is updated to $w_{k+1}$.
\vspace{4mm}
\section{Method}
The goal of our framework is {\em learning to fine-tune}. 
We propose MEta Loss TRansformer (MELTR), a novel auxiliary learning framework that adaptively combines auxiliary losses to assist fine-tuning on the target downstream task.
We formulate this as a bi-level optimization problem and 
present an efficient training procedure with Approximated Implicit Differentiation (AID) built on the Implicit Function Theorem (IFT).
Additionally, we introduce a regularization term to alleviate \textit{meta-overfitting} and learn a more effective combination of loss functions.

\begin{figure*}[t] 
    \centering
    \includegraphics[width=0.98\textwidth]{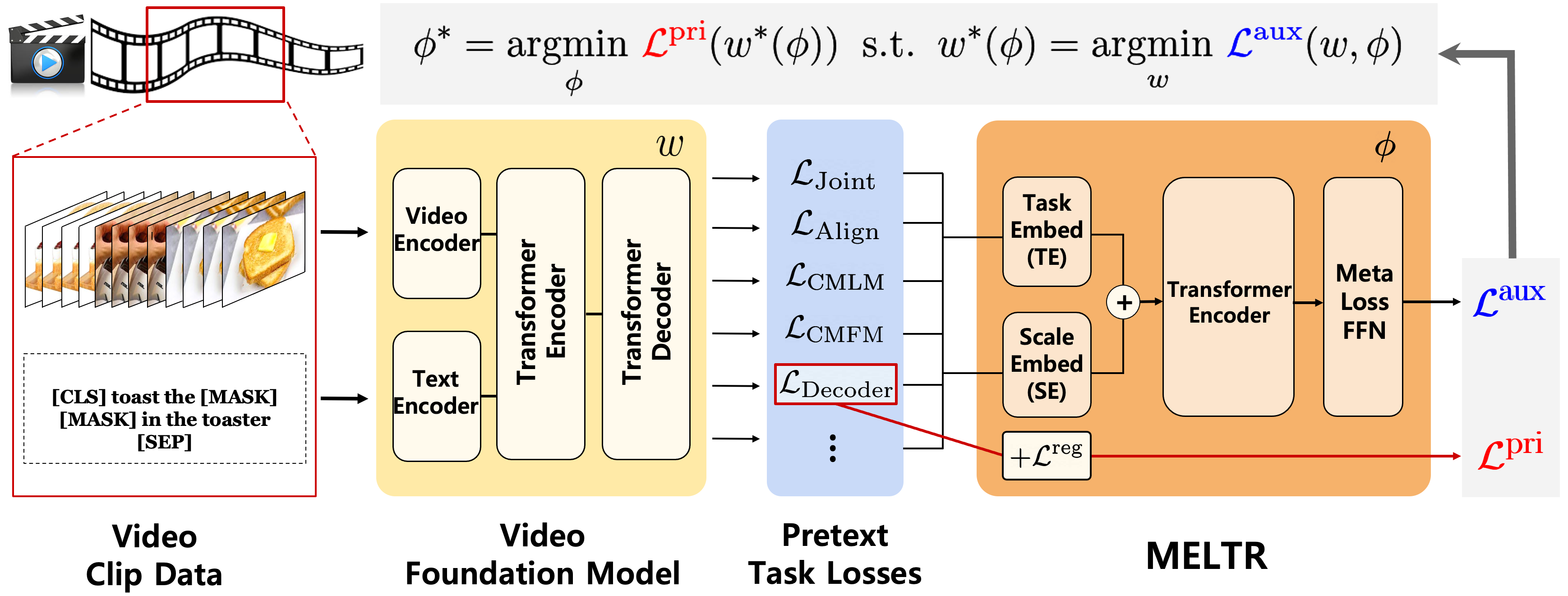}
    \caption{\textbf{Overall architecture.} 
    The Meta Loss Transformer (MELTR) is a plug-in module for meta auxiliary learning.
    The auxiliary pretext task losses derived from the video foundation model (\eg, UniVL~\cite{luo2020univl}) are input to MELTR, which is a transformer-based module that non-linearly aggregates the loss values from different tasks. The module is optimized to help learning of the primary task. 
    This figure illustrates the case when video captioning ($\mathcal{L}_\text{Decoder}$) is the primary task.}
    \label{fig:main}
\end{figure*}
\subsection{Meta Loss Transformer}
\label{subsec:MELTR}
Our framework generates a unified auxiliary loss function $\Laux$ by combining auxiliary losses $\Ljoint, \Lalign, \dots, \Ldecoder$.
In other words, our framework takes loss values from multiple auxiliary tasks and converts them to a new combined loss value as shown in Figure~\ref{fig:main}.
In order to leverage the relationship between primary and auxiliary tasks, we adopt the Transformer~\cite{vaswani2017attention} architecture.

Let $\mathcal{F}(\cdot ; w)$ denote a backbone foundation model parameterized by $w$. 
For $t$-th task, given input data $x$ and its label $y_t$, the loss value $\ell_t$ is defined as:
\begin{equation}
    \ell_t = \Lct(\mathcal{F}(x; w), y_t),
\end{equation}
where $\mathcal{L}_t$ is a loss function for $t$-th task.
With loss values $\Ell=[\ell_0, \ldots, \ell_T]$ from the primary task $t=0$ and auxiliary tasks $\{t=1, \ldots t=T \}$, our framework MELTR learns a unified auxiliary loss function defined as:
\begin{equation}
    \aux := \text{MELTR}(\Ell;\phi),
    \label{eq:MELTR}
\end{equation}
where $\text{MELTR}(\cdot;\phi)$ is a transformer-based neural network parameterized by $\phi$, which are meta-parameters in our meta-learning formulation.
In order to feed loss values, $\ell_0, \ldots, \ell_T$ to a Multi-head Self-attention layer, 
we transform a scalar loss value into the scale embedding (\textbf{SE}) and the task embedding (\textbf{TE}).
Each auxiliary loss value is first projected to a $d$-dimensional vector via $\textbf{SE}(\cdot)$, which is an MLP layer with a non-linear activation.
Similarly, we adopt a learnable embedding layer for \textbf{TE}, which plays the role of positional encodings.
Then, \textbf{SE} : $\bbR \rightarrow \bbR^{d}$ and \textbf{TE} : $\{0, \ldots, T \} \rightarrow \bbR^d$ are defined as:
\begin{equation}
    \textbf{SE}(\ell) := \text{MLP}(\ell) \text{, and } \textbf{TE}(t) := \text{Embedding}(t).
    \label{eq:sete}
\end{equation}
Then, the scale and task embeddings are summed to construct an input token. The input embeddings are self-attended and finally pooled to a scalar loss value, $\text{MELTR}(\Ell; \phi) \in \bbR$, by considering both the \textit{loss scale} and the \textit{task information}.
The overall architecture with the UniVL backbone is illustrated in Figure~\ref{fig:main}.

% Regularization
However, when {\em meta-data} (or a validation dataset) is small, meta-learning often suffers 
{\em meta-overfitting}~\cite{antoniou2018train,zintgraf2019fast}. 
In other words, meta-parameter $\phi$ may overfit to the primary task performance on small validation data.
To address this problem, we additionally introduce a regularization term $\mathcal{L}^\text{reg}$ given as:
\begin{equation}
    \mathcal{L}^\text{reg} = \left|\text{MELTR}(\Ell; \phi) - \sum_{t=0}^{T} \ell_t\right|.
    \label{eq:reg}
\end{equation}
This encourages the learned loss $\text{MELTR}(\Ell; \phi)$ to stay within a reasonable range.
Then, the primary task loss $\pri$, and the unified auxiliary loss $\aux$ are defined as follows:
\begin{equation}
    \pri = \mathcal{L}_0+\gamma\mathcal{L}^\text{reg}, \:\:\:
    \aux = \text{MELTR}(\Ell;\phi),
    \label{eq:loss}
\end{equation}
where $\gamma$ is a regularization strength and $\mathcal{L}_0$ is the original supervised loss for the target downstream task. For example, if $\Lalign$ is selected as the primary loss for the text-to-video retrieval task, then $\mathcal{L}_0 = \Lalign$ and all other tasks are considered as pretext tasks, \textit{i.e.}, $\Ell = [\Lalign, \Ljoint, \Lcmlm, \Lcmfm, \Ldecoder]$. 
Note that the primary loss itself is also included in the list of input loss functions.
\subsection{Objective function and optimization}
\label{subsec:objective}
MELTR learns how to fine-tune a model by non-linearly combining the auxiliary losses.
This can be viewed as hyperparameter optimization, which can be formulated as a bi-level optimization given as:
\begin{equation}
    \begin{split}
        & \phi^* = \argmin_{\phi} \:\pri(w^*(\phi)) \\
        \text{s.t.} \:\: & w^*(\phi) = \argmin_w \:\aux(w, \phi),
    \end{split}
    \label{eq:objective_function}
\end{equation}
where $\phi$ denotes the (meta) parameter of MELTR, and $w$ denotes the parameters of our backbone foundation model.
Then, we adopt one variant of the Approximate Implicit Differentiation (AID) scheme to optimize \eqref{eq:objective_function}.
Specifically, to optimize \eqref{eq:objective_function}, we first factorize the \textit{hypergradient}, which is the gradient of $\pri$ with respect to $\phi$ as $\nabla_\phi \pri = \nabla_w \pri \cdot \nabla_\phi w^*$, where
$\nabla_\phi w^* = -(\nabla_w^2 \aux)^{-1} \cdot \nabla_\phi\nabla_w \aux$ by the implicit function theorem (IFT).
Then, the hypergradient can be written as:
\begin{equation}
    \nabla_\phi \pri(w^*(\phi)) = -\nabla_w \pri \cdot \left(\nabla_w^2 \aux\right)^{-1} \cdot \nabla_\phi\nabla_w \aux.
    \label{eq:hypergrad}
\end{equation}
The evaluation of hypergradient entails the computation of the inverse of second-order derivatives.
In the literature~\cite{lorraine2020optimizing,navon2020auxiliary}, to accelerate the computation, the Neumann series is commonly adopted as:
\begin{equation}
    \left(\nabla_w^2 \aux\right)^{-1} = \lim_{i \rightarrow \infty} \sum_{j=0}^i \left(\mathrm{I} - \nabla_w^2 \aux\right)^j.
    \label{eq:neumann}
\end{equation}
In practice, the summation of an infinite series in \eqref{eq:neumann} is approximated by a finite sequence.
For instance, the number of iterations $i$ is usually truncated to a small integer (\textit{e.g.}, $i = 3$ in \cite{navon2020auxiliary}) in exchange for slight performance decay. 

\noindent However, this still requires considerable amount of time in the iterative computation of the Hessian matrix in \eqref{eq:neumann}.
We further simplify it by approximating the Hessian matrix in \eqref{eq:hypergrad} as the identity matrix $\mathrm{I}$. 
Then, our approximated gradient is given as follows:
\begin{equation}
    \nabla_\phi \pri(w^*(\phi)) \approx - \nabla_w \pri \cdot \nabla_\phi\nabla_w \aux.
    \label{eq:approximated}
\end{equation}
This completely removes the need for computation of the inverse Hessian matrix, which otherwise would have required a time complexity of $O(n^3)$. In our experiments, we observe that there is no significant degradation in terms of the performance of a fine-tuned model, see in Section~\ref{subsec:efficiency}.

Finally, with the approximated hypergradient \eqref{eq:approximated}, we utilize one variant of the AID scheme as an efficient optimization algorithm for MELTR.
We first optimize $w$ for $K$ steps by:
\begin{equation}
    w^{(k+1)} = w^{(k)} - \alpha \cdot \nabla_{w} \aux.
    \label{eq:update_w}
\end{equation}
After $K$ steps of \eqref{eq:update_w}, we then optimize for $\phi$ with:
\begin{equation}
    \begin{split}
        \phi^* & = \phi - \beta \cdot \nabla_\phi\pri(w^{(K)}(\phi)) \\
        & = \phi + \beta \cdot\left(\nabla_w\pri\cdot \nabla_\phi\nabla_w \aux\right),
    \end{split}
    \label{eq:update_phi}
\end{equation}
where $\alpha$ and $\beta$ are the learning rates of the backbone foundation model and MELTR, respectively.
The pseudo-code of our training scheme is provided in Algorithm~\ref{alg:main}.
% \begin{algorithm}[t!]
% \caption{L2F Algorithm with MTN}
% \label{alg:main}
% \textbf{Inputs:} $w$, $\phi$ \\
% \textbf{Parameters:} learning rate $\alpha$, $\beta$ \\
% % \vspace{-10pt}
% \begin{algorithmic}[1]
%     \For{$e = 1$ {\bfseries to} Epochs}
%         \For{$k = 1$ {\bfseries to} $K_1$}
%             \State $\ell_t = f_t(x^k, y^k; w) \:\:\: \forall t \in [0, T]$
%             \State $\aux \leftarrow MTN(\boldsymbol{\ell}; \phi)$
%             \State $w \leftarrow w - \alpha \cdot \nabla_w \aux \Big|_{w, \phi}$
%         \EndFor
%         \For{$k = 1$ {\bfseries to} $K_2$}
%             \State $\ell_t = f_t(x^k, y^k; w) \:\:\: \forall t \in [0, T]$
%             \State $\pri, \: \aux \leftarrow \ell_0, \: MTN(\boldsymbol{\ell}; \phi)$
%              \State $\phi \leftarrow \phi + \beta \cdot \nabla_w \pri \Big|_{w, \phi} \cdot \nabla_\phi\nabla_w \aux \Big|_{w, \phi}$
%         \EndFor
%     \EndFor
% \end{algorithmic}
% \textbf{return} $w$

% \end{algorithm}

\begin{algorithm}[t!]
\caption{MELTR optimization algorithm}\label{alg:opt}
\label{alg:main}
\textbf{Inputs:} $w$, $\phi$ \\
\textbf{Parameters:} learning rate ($\alpha$, $\beta$), regularization coefficient $\gamma$, inner iter $K$ \\
% \vspace{-10pt}
\begin{algorithmic}[1]
    % \For{$e = 1$ {\bfseries to} Epochs}
    \vspace{-3mm}
    \While{not converged}
        \For{$k = 1$ {\bfseries to} $K$}
            % \State $\ell_t = f_t(x^k, y^k_t; w) \:\:\: \forall t \in [0, T]$
            \State $\ell_t = \mathcal{L}_t(\mathcal{F}(x; w), y_t) \:\:\: \forall t \in [0, T]$
            \State $\aux \leftarrow \text{MELTR}(\boldsymbol{\ell}; \phi)$
            \State $w \leftarrow w - \alpha \cdot \nabla_w \aux \Big|_{w, \phi}$
        \EndFor
        \State $\ell_t = \mathcal{L}_t(\mathcal{F}(x; w), y_t) \:\:\: \forall t \in [0, T]$
        \State $\aux \leftarrow \text{MELTR}(\boldsymbol{\ell}; \phi)$
        \State $\pri \leftarrow \ell_0+ \gamma \cdot \left|\text{MELTR}(\boldsymbol{\ell}; \phi) - \sum_t \ell_t\right|$
        \State $\phi \leftarrow \phi + \beta \cdot \nabla_w \pri \Big|_{w, \phi} \cdot \nabla_\phi\nabla_w \aux \Big|_{w, \phi}$
    \EndWhile
    \State \textbf{return} $w$
    \vspace{1mm}
\end{algorithmic}
\end{algorithm}
\section{Experiments}
To verify the effectiveness of our method, we apply it to multiple video foundation models (UniVL~\cite{luo2020univl}, Violet~\cite{fu2021violet}, All-in-one~\cite{wang2022all}), and evaluate them on four downstream tasks: text-to-video retrieval, video question answering, video captioning, and multimodal sentiment analysis. 
For the tasks, we use five benchmark datasets: YouCook2~\cite{zhou2018towards}, MSRVTT~\cite{xu2016msr}, TGIF-QA~\cite{jang2017tgif}, MSVD-QA~\cite{xu2017video}, CMU-MOSI~\cite{zadeh2016mosi}.
We conduct experiments and analyze the results to answer the following research questions:\\
\textbf{Q1.} Does the learned combination of auxiliary losses benefit the primary task?\\
\textbf{Q2.} What does MELTR learn from auxiliary learning?\\
\textbf{Q3.} Is the proposed optimization method efficient for MELTR?

\noindent \textbf{Datasets.} 
For video retrieval, we use YouCook2 and MSRVTT.
For video question answering, TGIF-QA and MSVD-QA datasets are used, and YouCook2 and MSRVTT are used for video captioning.
Finally, we use CMU-MOSI for multi-modal sentiment analysis.
Further dataset details are provided in the supplement.

\noindent \textbf{Implementation details.}
MELTR is adapted to UniVL~\cite{luo2020univl}, Violet~\cite{fu2021violet}, and All-in-one~\cite{wang2022all} for main experiments and we conduct ablation studies and qualitative analyses on UniVL.
As for UniVL, we use five auxiliary loss functions ($\Ljoint$, $\Lalign$, $\Lcmlm$, $\Lcmfm$, and $\Ldecoder$), which were introduced in Section~\ref{subsec:univl}.
We additionally adopt three advanced auxiliary loss functions, $\Lmjoint$, $\Lmalign$, and $\Lmdecoder$, which leverage masked language and masked visual features obtained by converting some of the language or visual tokens into [MASK] or random tokens, along with the five objectives described above.
For text-to-video retrieval and video captioning, $\Lalign$ and $\Ldecoder$ are used as the primary loss functions, respectively.
Further implementation details including Violet and All-in-one are in the supplement.

\begin{table}[t!]
    \centering
    \caption{\textbf{Text-to-Video retrieval on YouCook2.} 
    UniVL-Joint and UniVL-Align denote the model fine-tuned with the $\Ljoint$ and $\Lalign$, respectively.
    MELTR is applied to the UniVL-Align. 
    MELTR$^-$ refers to MELTR without the regularization term $\mathcal{L}^\text{reg}$.
    }
    \begin{adjustbox}{width=\linewidth}
    \begin{tabular}{l|c c c c}
        \toprule
        \multicolumn{1}{c|}{\textbf{Models}} & R@1$\uparrow$ & R@5$\uparrow$ & R@10$\uparrow$ & MedR$\downarrow$ \\
        \midrule
        \midrule
        HGLMM-FV-CCA~\cite{klein2015associating}     & 4.6  & 21.6 & 14.3 & 75 \\
        HowTo100M~\cite{miech2019howto100m}          & 8.2  & 35.3 & 24.5 & 24 \\
        ActBERT~\cite{zhu2020actbert}                & 9.6  & 26.7 & 38.0 & 19  \\
        MIL-NCE~\cite{miech2020end}                  & 15.1 & 38.0 & 51.2 & 10  \\
        COOT~\cite{ging2020coot}                     & 16.7 & 40.2 & 25.3 & 9  \\
        TACo~\cite{yang2021taco}                     & 29.6 & 59.7 & 72.7 & 9  \\
        VideoCLIP~\cite{xu2021videoclip}             & 32.2 & 62.6 & \textbf{75.0} & - \\
        \midrule
        UniVL-Joint~\cite{luo2020univl}              & 22.2 & 52.2 & 66.2 & 5  \\
        UniVL-Align~\cite{luo2020univl}              & 28.9 & 57.6 & 70.0 & 4  \\
        \rowcolor{cyan!25}
        UniVL + \textbf{MELTR}$^-$                   & 33.4 & 62.5 & 73.3 & \textbf{3} \\
        \rowcolor{cyan!25}
        UniVL + \textbf{MELTR}                       & \textbf{33.7} & \textbf{63.1} & 74.8 & \textbf{3} \\
        \bottomrule
    \end{tabular}
    \end{adjustbox}
    \label{tab:cook_ret}
\end{table}
\begin{table}[t!]
    \centering
    \setlength{\tabcolsep}{2.0pt}
    \caption{\textbf{Text-to-Video retrieval on MSRVTT.} 
    }
    \begin{adjustbox}{width=\linewidth}
    \begin{tabular}{l|c c c c|c c c c}
        \toprule
        \multicolumn{1}{c}{\textbf{Models}} & \multicolumn{4}{|c}{\textbf{MSRVTT-7k}} & \multicolumn{4}{|c}{\textbf{MSRVTT-9k}} \\
        & R@1$\uparrow$ & R@5$\uparrow$ & R@10$\uparrow$ & MedR$\downarrow$ & R@1$\uparrow$ & R@5$\uparrow$ & R@10$\uparrow$ & MedR$\downarrow$ \\
        \midrule
        \midrule
        MIL-NCE~\cite{miech2020end}        & 9.9  & 24.0 & 32.4 & 29.5 & - & - & - & - \\
        JSFusion~\cite{yu2018joint}        & 10.2 & 31.2 & 43.2 & 13 & - & - & - & - \\
        HowTo100M~\cite{miech2019howto100m}& 14.9 & 40.2 & 52.8 & 9 & - & - & - & - \\
        HERO~\cite{li2020hero}             & 16.8 & 43.4 & 57.7 & - & - & - & - & - \\
        ClipBERT~\cite{lei2021clipbert}    & 22.2 & 46.8 & 59.9 & 6 & - & - & - & - \\
        MMT~\cite{gabeur2020multi}         & -    & -    & -    & - & 26.6 & 57.1 & 69.6 & 4 \\
        T2VLAD~\cite{wang2021t2vlad}       & -    & -    & -    & - & 29.5 & 59.0 & 70.1 & 4 \\
        TACo~\cite{yang2021taco}           & 19.2 & 44.7 & 57.2 & 7 & 28.4 & 57.8 & 71.2 & 4 \\
        VideoCLIP~\cite{xu2021videoclip}   & -    & -    & -    & - & 30.9 & 55.4 & 66.8 & - \\
        Frozen~\cite{bain2021frozen}       & -    & -    & -    & - & 32.5 & 61.5 & 71.2 & 3 \\
        \midrule
        UniVL-Joint~\cite{luo2020univl}    & 20.6 & 49.1 & 62.9 & 6 & 27.2 & \textbf{55.7} & \textbf{68.7} & \textbf{4}  \\
        UniVL-Align~\cite{luo2020univl}    & 21.2 & 49.6 & 63.1 & 6 & - & - & - & - \\
        \rowcolor{cyan!25}
        UniVL + \textbf{MELTR}  & \textbf{28.5} & \textbf{55.5} & \textbf{67.6} & \textbf{4} & \textbf{31.1} & \textbf{55.7} & 68.3 & \textbf{4} \\
        \midrule
        Violet~\cite{fu2021violet}         & 31.7 & 60.1 & 74.6 & \textbf{3} & 34.5 & 63.0 & 73.4 & - \\
        \rowcolor{cyan!25}
        Violet + \textbf{MELTR}            & \textbf{33.6} & \textbf{63.7} & \textbf{77.8} & \textbf{3} & \textbf{35.5} & \textbf{67.2} & \textbf{78.4} & \textbf{3} \\
        \midrule
        All-in-one~\cite{wang2022all}      & 34.4 & 65.4 & 75.8 & - & 37.9 & 68.1 & 77.1 & - \\
        \rowcolor{cyan!25}
        All-in-one + \textbf{MELTR}        & \textbf{38.6} & \textbf{74.4} & \textbf{84.7} & - & \textbf{41.3} & \textbf{73.5} & \textbf{82.5} & - \\
        \bottomrule
    \end{tabular}
    \end{adjustbox}
    \label{tab:vtt_ret}
\end{table}
\subsection{Evaluation on downstream tasks}
\label{subsec:downstream}

We here answer \textbf{Q1} (the effectiveness of MELTR) by applying our framework to fine-tune the pretrained foundation models on various downstream tasks: text-to-video retrieval, video question answering, video captioning, and multi-modal sentiment analysis.

\noindent \textbf{Text-to-Video retrieval.} 
We evaluate text-to-video retrieval task performance on YouCook2 and MSRVTT.
Table~\ref{tab:cook_ret} shows that our method outperforms all the baseline models by plugging MELTR in the standard UniVL.
Specifically, the R@1 is improved by 4.8\% compared to UniVL, and 1.5\% compared to the previous SOTA, VideoCLIP~\cite{yang2021taco}.
Also, METLR with the regularization term $\mathcal{L}^\text{reg}$ improves all the performance metrics compared to MELTR$^-$, which does not use $\mathcal{L}^\text{reg}$.
This optional regularization term confines the loss value to a reasonable bound, which prevents meta-overfitting.

\begin{table}[t!]
    \centering
    \caption{\textbf{Video question answering on TGIF-QA and MSVD-QA.}
    }
    \begin{adjustbox}{width=\linewidth}
    \begin{tabular}{l|c c c|c}
        \toprule
        \multicolumn{1}{c|}{\textbf{Models}} & \multicolumn{3}{c|}{\textbf{TGIF-QA}} & \textbf{MSVD-QA} \\ 
        & Action & Transition & Frame & \\
        \midrule
        \midrule
        HME~\cite{fan2019heterogeneous} & 73.9 & 77.8 & 53.8 & 33.7 \\
        HCRN~\cite{le2020hierarchical} & 75.0 & 81.4 & 55.9 & 36.1 \\
        QueST~\cite{jiang2020divide} & 75.9 & 81.0 & 59.7 & 36.1 \\
        ClipBERT~\cite{lei2021less} & 82.9 & 87.5 & 59.4 & - \\
        \midrule
        Violet~\cite{fu2021violet} & 92.5 & 95.7 & 62.3 & 47.9 \\
        \rowcolor{cyan!25}
        Violet + \textbf{MELTR} & \textbf{95.4} & \textbf{97.5} & \textbf{63.4} & \textbf{51.7} \\
        \bottomrule
    \end{tabular}
    \end{adjustbox}
    \label{tab:qa}
\end{table}
\begin{table}[t!]
    \centering
    \setlength{\tabcolsep}{3.5pt}
    \caption{\textbf{Video captioning on YouCook2.} 
    V refers to the video-only input setting, and V+T the multi-modal setting with video and transcript inputs.
    }
    \begin{adjustbox}{width=\linewidth}
    \begin{tabular}{l|c|c c c c c}
        \toprule
        \multicolumn{1}{c|}{\textbf{Models}} & Modality & BLEU-3 & BLEU-4 & METEOR & ROUGE-L & CIDEr  \\
        \midrule
        \midrule
        EMT~\cite{zhou2018end}                          & V   & 7.53 & 4.38 & 11.55 & 27.44 & 38 \\
        CBT~\cite{sun2019learning}                      & V   &  -   & 5.12 & 12.97 & 30.44 & 64 \\
        ActBERT~\cite{zhu2020actbert}                   & V   & 8.66 & 5.41 & 13.30 & 30.56 & 65 \\
        VideoBERT~\cite{sun2019videobert}               & V   & 6.33 & 3.81 & 10.81 & 27.14 & 47 \\
        COOT~\cite{ging2020coot}                        & V   & 17.97& 11.30 & 19.85 & 37.94 & 57 \\
        VideoBERT~\cite{sun2019videobert}               & V+T & 7.59 & 4.33 & 11.94 & 28.80 & 55 \\
        DPC~\cite{shi2019dense}                         & V+T & 7.60 & 2.76 & 18.08 & -     & -    \\
        AT+Video~\cite{hessel2019case}                  & V+T &  -   & 9.01 & 17.77 & 36.65 & 112 \\
        \midrule
        UniVL~\cite{luo2020univl}                       & V   & 16.46 & 11.17 & 17.57 & 40.09 & 127 \\
        \rowcolor{cyan!25}
        UniVL + \textbf{MELTR}                          & V   & 17.35 & 11.98 & 18.19 & 41.28 & 138 \\
        UniVL~\cite{luo2020univl}                       & V+T & 23.87 & 17.35 & 22.35 & 46.52 & 181 \\
        \rowcolor{cyan!25}
        UniVL + \textbf{MELTR}                          & V+T & \textbf{24.12} & \textbf{17.92} & \textbf{22.56} & \textbf{47.04} & \textbf{190} \\
        \bottomrule
    \end{tabular}
    \end{adjustbox}
    \label{tab:cook_cap}
\end{table}

In Table~\ref{tab:vtt_ret}, our model outperforms all the baselines including foundation models and task-specific methods in all the retrieval metrics.
Specifically, MELTR improved three baseline foundation models: UniVL, Violet, and All-in-one.
For each model, R@1 is improved by a margin of 7.3\%, 1.9\%, and 4.2\% on MSRVTT-7k by plugging in MELTR.
The R@1 is also improved by a large margin of 4.8\%, 7.3\%, and 3.9\% respectively on YouCook2, MSRVTT-7k, and MSRVTT-9k as well, compared to the standard UniVL variants denoted UniVL-Joint or UniVL-Align.

\noindent \textbf{Video question answering.}
We experiment video question answering on TGIF-QA and MSVD-QA in Table~\ref{tab:qa}.
Plugging MELTR in the foundation model outperforms all the baselines.
Especially in MSVD-QA, MELTR obtains a large margin of 3.8\% improvement over the standard Violet.

\noindent \textbf{Video captioning.}
In Table~\ref{tab:cook_cap} and Table~\ref{tab:vtt_cap}, we evaluate video captioning task performance on YouCook2 and MSRVTT.
In the case of YouCook2, we conduct experiments on the `video-input-only' setting and additionally experiment on `video + text (transcript)' input, following previous works.
MELTR outperforms all the baseline models, in terms of all metrics.
In the case of MSRVTT, the performance of MELTR significantly improves BLEU scores, which are the major performance metric, BLEU-4.

\begin{table}[t!]
    \centering
    \setlength{\tabcolsep}{3.5pt}
    \caption{\textbf{Video captioning on MSRVTT-full.} 
    $^*$ refers to the experimental results reported in the official github.
    }
    \begin{adjustbox}{width=\linewidth}
    \begin{tabular}{l|c|c c c c c}
        \toprule
        Models & Modality & BLEU-3 & BLEU-4 & METEOR & ROUGE-L & CIDEr  \\
        \midrule
        \midrule
        PickNet~\cite{chen2018less}                     & V   & - & 35.6 & 26.8 & 58.2 & 41.0 \\
        PickNet~\cite{chen2018less}                     & V+T & - & 38.9 & 27.2 & 59.5 & 42.1 \\
        MARN~\cite{pei2019marn}                         & V   & - & 40.4 & 28.1 & 60.7 & 47.1 \\
        SibNet~\cite{liu2020sibnet}                     & V   & - & 40.9 & 27.5 & 60.2 & 47.5 \\
        OA-BTG~\cite{zhang2019object}                   & V   & - & 41.4 & 28.2 & -    & 46.9 \\
        POS-VCT~\cite{hou2019joint}                     & V   & - & 42.3 & \textbf{29.7} & \textbf{62.8} & 49.1 \\
        ORG-TRL~\cite{zhang2020object}                  & V   & - & 43.6 & 28.8 & 62.1 & 50.9 \\
        \midrule
        UniVL$^*$~\cite{luo2020univl}                   & V   & 53.42 & 41.79 & 28.94 & 60.78 & 50.04 \\
        \rowcolor{cyan!25}
        UniVL + \textbf{MELTR}                          & V   & \textbf{55.88} & \textbf{44.17} & 29.26 & 62.35 & \textbf{52.77} \\
        \bottomrule
    \end{tabular}
    \end{adjustbox}
    \label{tab:vtt_cap}
\end{table}
\begin{figure}[t] 
    \centering
    \includegraphics[width=\linewidth]{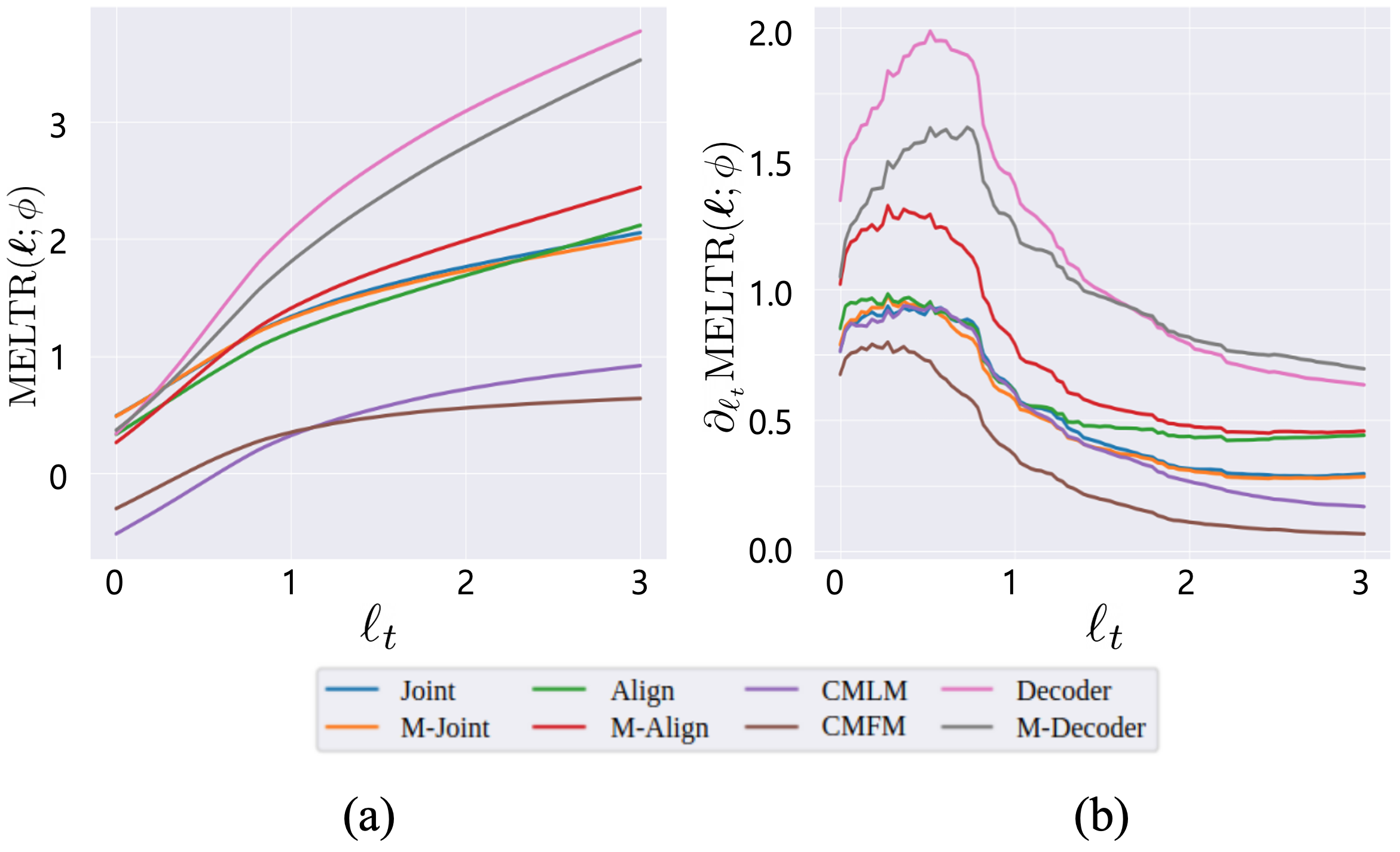}
    \caption{\textbf{$\text{MELTR}(\boldsymbol{\ell};\phi)$ and $\partial_{\ell_t} \text{MELTR}(\boldsymbol{\ell};\phi).$}
    The combined loss MELTR$(\boldsymbol{\ell};\phi)$ in (a) and the partial derivative of MELTR$(\boldsymbol{\ell};\phi)$ with respect to $\ell_t$ in (b) are visualized by changing each pretext task loss from 0 to 3, while other losses are fixed to their average values. 
    In the video captioning task, the Decoder loss and the Masked-Decoder loss rendered the highest MELTR$(\boldsymbol{\ell};\phi$) and $\partial_{\ell_t}$MELTR$(\boldsymbol{\ell};\phi)$ values overall.
    }
    \label{fig:scale}
    \vspace{-3mm}
\end{figure}

\noindent \textbf{Multi-modal sentiment analysis.}
We also experiment the multi-modal sentiment analysis task on CMU-MOSI.
Table~\ref{tab:sentiment} shows that MELTR surpasses all the baselines.
These experimental results indicate that MELTR is successful in adaptively combining the auxiliary losses across backbone model architectures on various tasks.
\subsection{Analysis on MELTR}

We discuss \textbf{Q2} by analyzing how MELTR combines the losses.
We first analyze the non-linear relationship between the input and output loss values of MELTR, and examine how MELTR adaptively re-weights the auxiliary tasks.
Note, we use MELTR trained for the video captioning task on YouCook2 for these analyses, and abbreviate $\partial_{\ell_t} \text{MELTR}(\Ell;\phi):=\frac{\partial}{\partial \ell_t} \text{MELTR}(\Ell;\phi)$ hereafter.

\begin{table}[t!]
    \centering
    \small
    \caption{\textbf{Multimodal sentiment analysis on CMU-MOSI.}
    BA, F1, MAE, and Corr are binary accuracy, F1 score, mean absolute error, and Pearson correlation coefficient, respectively.
    }
    \begin{tabular}{l|c c c c}
        \toprule
        Models & BA$\uparrow$ & F1$\uparrow$ & MAE$\downarrow$ & Corr$\uparrow$ \\
        \midrule
        \midrule 
        MulT                      & 83.0 & 82.8 & 0.870 & 0.698 \\
        FMT                       & 83.5 & 83.5 & 0.837 & 0.744 \\
        \midrule
        UniVL                     & 84.6 & 84.6 & 0.781 & 0.767 \\
        \rowcolor{cyan!25}
        UniVL + \textbf{MELTR}    & \textbf{85.3} & \textbf{85.4} & \textbf{0.759} & \textbf{0.789} \\
        \bottomrule
    \end{tabular}
    \label{tab:sentiment}
\end{table}
\begin{figure}[t] 
    \centering
    \includegraphics[width=\linewidth]{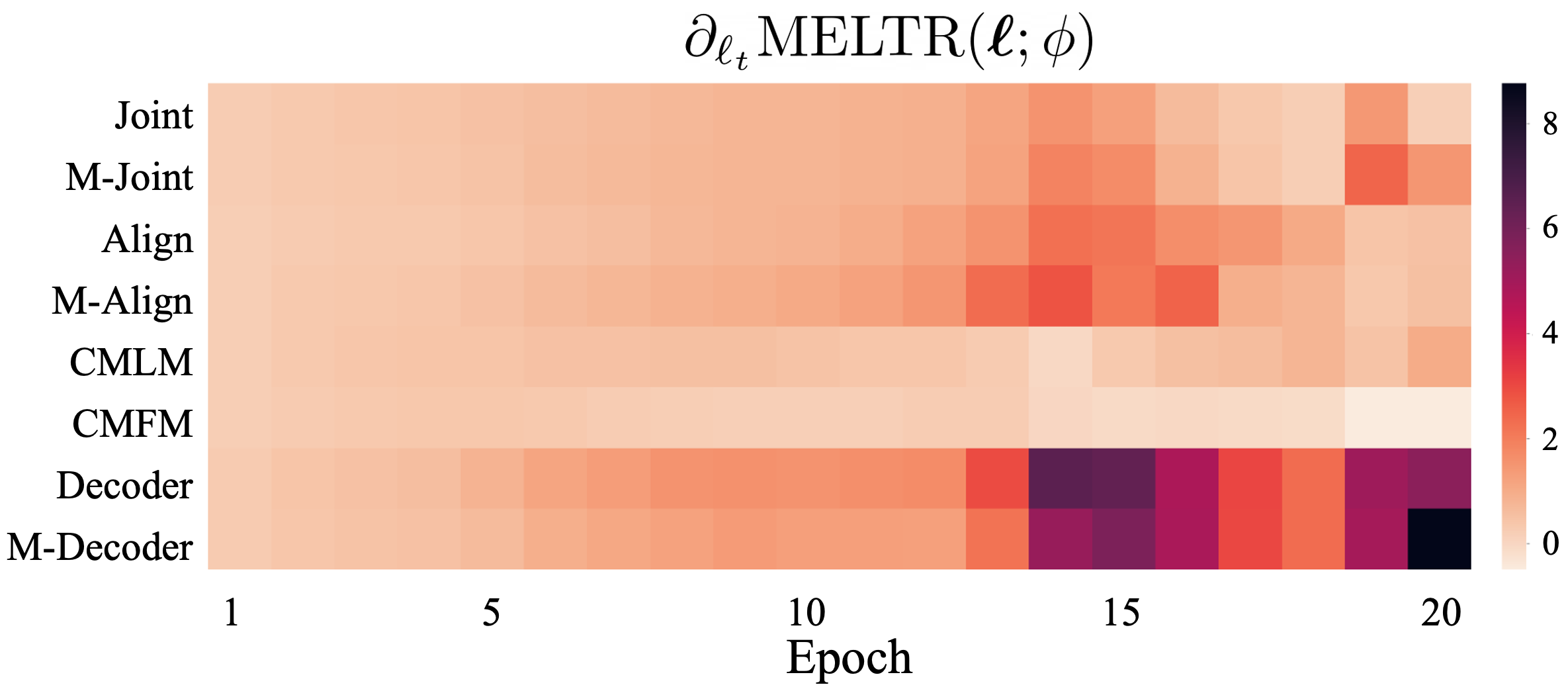}
    \caption{
    \textbf{Illustration of $\partial_{\ell_t}$MELTR.} 
    The $\partial_{\ell_t}$MELTR$(\boldsymbol{\ell};\phi$) values for the pretext task losses are plotted for each epoch, when training for video captioning. 
    The gradients are computed by taking the average for each data sample per epoch. 
    In the first few epochs, all task losses possess a similar scale of gradients. 
    As training continues, the most relevant losses, Decoder and M-Decoder, receives larger gradients, while the least relevant, CMFM, has the smallest gradient value.}
    \label{fig:epoch}
    \vspace{-3mm}
\end{figure}

\noindent \textbf{Non-linear loss transformation.}
Figure~\ref{fig:scale}(a) shows that all auxiliary losses are positively correlated with the output loss. 
We can also observe that the output and input are non-linearly correlated.
On the other hand, Figure~\ref{fig:scale}(b) shows that $\partial_{\ell_t} \text{MELTR}(\Ell;\phi)$ have relatively higher values around $\ell_t = 0.5$ and the gradient becomes smaller as $\ell_t$ increases.
This indicates that $\text{MELTR}$ guides the learner to focus on reasonably challenging samples and if the loss is too large, it becomes less sensitive (\ie, too large an input loss is interpreted as noise and it tends to be downweighted).
Also, given the primary task loss $\Ldecoder$ for video captioning, the MELTR is more sensitive to the change of $\Ldecoder$ and $\Lmdecoder$ rather than $\Lcmfm$.
In other words, MELTR learned that text generation task losses ($\Ldecoder$ and $\Lmdecoder$) are more relevant to the video captioning than masked frame generation ($\Lcmfm$).

\begin{table*}[t!]
    \centering
    \caption{\textbf{Comparison of various multi-task learning schemes.}
    We compare MELTR with manually designed five multi-task learning schemes:
    (A) adopts only $\Ldecoder$,
    (B) adopts both $\Ldecoder$ and $\Lmdecoder$ which are useful for video captioning based on our observation,
    (C) fixes all the coefficients to 1,
    (D) drops only $\Lcmfm$ which is useless for video captioning from (C) based on our observation,
    and (E) re-weights the loss coefficients based on task importance, contrary to (D).
    }
    \begin{adjustbox}{width=\linewidth}
    \begin{tabular}{c|c c c c c c c c|c c c c c}
        \toprule
        \textbf{Models} & \multicolumn{8}{c|}{\textbf{Coefficient of each task}} & \multicolumn{5}{c}{\textbf{Video captioning on YouCook2}} \\
        & $\Ljoint$ & $\Lmjoint$ & $\Lalign$ & $\Lmalign$ & $\Lcmlm$ & $\Lcmfm$ & $\Ldecoder$ & $\Lmdecoder$ & BLEU-3 & BLEU-4 & METEOR & ROUGE-L & CIDEr  \\
        \midrule
        \midrule
        (A) & 0 & 0 & 0 & 0 & 0 & 0 & 1 & 0 & 22.79 & 16.54 & 21.73 & 45.85 & 1.78 \\
        (B) & 0 & 0 & 0 & 0 & 0 & 0 & 1 & 1 & 23.42 & 17.14 & 22.27 & 46.65 & 1.85 \\
        (C) & 1 & 1 & 1 & 1 & 1 & 1 & 1 & 1 & 21.72 & 15.93 & 20.89 & 45.16 & 1.79 \\
        (D) & 1 & 1 & 1 & 1 & 1 & 0 & 1 & 1 & 21.99 & 16.10 & 21.09 & 45.35 & 1.85 \\
        (E) & 1 & 1 & 1 & 1 & 1 & 0 & 8 & 8 & 23.31 & 17.23 & 21.98 & 46.26 & 1.85 \\
        \midrule
        \textbf{MELTR} & \multicolumn{8}{c|}{ADAPTIVE} & \textbf{24.12} & \textbf{17.92} & \textbf{22.56} & \textbf{47.04} & \textbf{1.90} \\
        \bottomrule
    \end{tabular}
    \end{adjustbox}
    \label{tab:manual}
    \vspace{-2mm}
\end{table*}

\noindent \textbf{Adaptive task re-weighting.}
As an extension of the above observation, we visualize $\partial_{\ell_t} \text{MELTR}(\Ell;\phi)$ for each epoch in Figure~\ref{fig:epoch}.
At the beginning of training, MELTR equally takes into account all the auxiliary tasks.
As training proceeds, MELTR evaluates that $\Ldecoder$ and $\Lmdecoder$ are effective for the primary loss $\Ldecoder$, while $\Lcmfm$ is relatively less beneficial, if not harmful.
This is consistent with our observation in Figure~\ref{fig:scale} since $\Ldecoder$ and $\Lmdecoder$ mainly conduct the text generation task while $\Lcmfm$ is for masked frame generation.

In Table~\ref{tab:manual}, we also compare MELTR with five manually designed multi-task learning schemes, each combining the task losses with different linear coefficients.
First, by comparing (A) and (B), an auxiliary loss $\Lmdecoder$ assists learning of the video captioning task.
However, (A) and (C) demonstrate that multi-task learning is not always beneficial, and it sometimes hinders fine-tuning if the auxiliary tasks include harmful task losses.
By dropping $\Lcmfm$ from (C), the model performance is slightly improved in (D).
Interestingly, this matches our observation that $\Lcmfm$ is disadvantageous for video captioning (Figure~\ref{fig:epoch}).
Furthermore, (E) outperforms (D), implying that re-weighting among the auxiliary tasks can be beneficial for multi-task learning.
Finally, our MELTR surpasses all the multi-task learning schemes above.
These experimental results indicate that MELTR effectively learns to fine-tune by adaptively re-weighting the auxiliary tasks, compared to the coarse and heuristically designed multi-task learning schemes.
\subsection{Efficient optimization algorithm}
\label{subsec:efficiency}

We also discuss the optimization algorithms for training MELTR to answer the last question \textbf{Q3}.
We compare various bi-level optimization algorithms based on ITD or AID schemes.
To analyze the efficiency, we measure the latency of an epoch and the performance on the text-to-video retrieval task with MSRVTT-7k.
We identically use the MELTR module as the loss combining network with all optimization algorithms for fair comparisons.
We also compare with the multi-task learning setting where the model is fine-tuned with a linearly summed loss.

In Table~\ref{tab:efficiency}, the multi-task learning (MTL) method is faster than meta-learning-based algorithms (denoted by `\textbf{MELTR} + $\alpha$')
since MTL is formulated as a uni-level optimization problem.
We observe that all MELTR with various bi-level optimization schemes outperform a Multi-task Learning in terms of the target task performance R@1.
Among the bi-level optimization schemes, our training scheme denoted as \textbf{MELTR + AID-FP-Lite$^{\dagger}$} introduces only 4.9\% overhead in training time than multi-task learning, while improving performance by 2.4\%.
This is a significant improvement considering that Meta-Weight Net (ITD) takes longer than twice the time required by Multi-task Learning and AID-FP-Lite.
Our optimization in Algorithm~\ref{alg:opt}, which approximates $\nabla^2_w \aux$ in \eqref{eq:hypergrad} with the identity matrix $\mathrm{I}$, is the fastest bi-level optimization scheme in Table~\ref{tab:efficiency} while achieving a strong R@1 performance. 
\begin{table}[t!]
    \centering
    \caption{
    \textbf{Efficiency comparison of optimization algorithms.}
    R@1 scores evaluated on MSRVTT-7k for video retrieval are recorded.
    Multi-task learning simultaneously trains all tasks with even loss weights. 
    CG and FP are abbreviations of conjugate gradient and fixed-point optimization. 
    In terms of time costs, average training time per epoch is reported. 
    $^\dagger$ refers to our optimization algorithm which approximates $\nabla^2_w \aux$ as the identity matrix $\mathrm{I}$.}
    \begin{adjustbox}{width=\linewidth}
    \begin{tabular}{l |c| c  c}
        \toprule
        \textbf{Method}  & \textbf{Opt. Scheme}  & \textbf{R@1} &  \textbf{Time} \\
        \midrule
        \midrule
        Multi-task Learning   & 
        - &  
        26.1 \scriptsize(+0.0)    & 
        547 \scriptsize(+0.0\%) \\
        
        \textbf{MELTR} + Meta-Weight Net~\cite{shu2019meta}  & 
        ITD &  
        27.3 \scriptsize(\textcolor{red}{+1.2})  & 
        1,296 \scriptsize(\textcolor{red}{+136.9\%}) \\ 
        
        \textbf{MELTR} + StocBIO~\cite{ji2021bilevel} & 
        N/A  &  
        26.8 \scriptsize(\textcolor{red}{+0.7})   &   
        686 \scriptsize(\textcolor{red}{+25.4\%})\\
        
        \textbf{MELTR} + CG & 
        AID-CG &  
        28.0 \scriptsize(\textcolor{red}{+1.9})   &   
        624 \scriptsize(\textcolor{red}{+14.1\%})\\
        
        \textbf{MELTR} + AuxiLearn~\cite{navon2020auxiliary} &  
        AID-FP    &  
        27.9 \scriptsize(\textcolor{red}{+1.8})    &
        638 \scriptsize(\textcolor{red}{+16.6\%})      \\
        
        \textbf{MELTR} + \textbf{AID-FP-Lite}$^\dagger$ & 
        AID-FP &  
        28.5 \scriptsize(\textcolor{red}{+2.4})   &   
        574 \scriptsize(\textcolor{red}{+4.9\%})\\
        \bottomrule
    \end{tabular}
    \end{adjustbox}
    \label{tab:efficiency}
    \vspace{-3mm}
\end{table}
\section{Conclusion}

We proposed Meta Loss Transformer (MELTR), an auxiliary learning framework that learns to fine-tune video foundation models.
MELTR learns to integrate various pretext task losses into one loss function to boost the performance of the target downstream task.
Our qualitative analysis demonstrates that MELTR improves the performance of the primary task by considering the type of task and the scale of the loss value.
The proposed training procedure built on AID-FP-Lite with a simple approximation of the inverse Hessian matrix achieved the efficiency without a significant performance loss.
By plugging MELTR into various foundation models, our method outperformed state-of-the-art video foundation models as well as task-specific models on a wide range of downstream tasks.

\noindent \textbf{Acknowledgments.}
This work was partly supported by ICT Creative Consilience program (IITP-2023-2020-0-01819) supervised by the IITP; Electronics and Telecommunications Research Institute (ETRI) grant funded by the Korean government (23ZS1200, Fundamental Technology Research for Human-Centric Autonomous Intelligent Systems); and KaKaoBrain corporation.

\appendix

\section*{\Large Appendix}
\section{Implementation Details}

\subsection{Backbone Foundation Models}
\noindent \textbf{UniVL~\cite{luo2020univl}.}
Our implementation is based on the official code of UniVL~\cite{univl_github} pretrained on the HowTo100M dataset~\cite{miech2019howto100m}.
As in the main paper, we use eight auxiliary loss functions: $\Ljoint$, $\Lmjoint$, $\Lalign$, $\Lmalign$, $\Lcmlm$, $\Lcmfm$, $\Ldecoder$, and $\Lmdecoder$.
For the primary losses for text-to-video retrieval and video captioning tasks are $\Lalign$ and $\Ldecoder$, respectively.
$\Lalign$ is also used as the primary loss function for multi-modal sentiment analysis.

\noindent \textbf{Violet~\cite{fu2021violet}.}
We implement MELTR based on the official Violet github~\cite{violet_github} pretrained on the YT-Temporal 180M~\cite{zellers2021merlot}, WebVid~\cite{bain2021frozen}, and CC3M~\cite{sharma2018conceptual}.
For text-to-video retrieval, we adopt three auxiliary losses: video-text matching loss, masked text modeling loss, and masked visual-token modeling. 
We use the former one as the primary task loss.
We use additional classification loss for video question answering.

\noindent \textbf{All-in-one~\cite{wang2022all}.}
Our implementation for All-in-one is based on \cite{all_github} and it is pretrained on WebVid~\cite{bain2021frozen}, YT-Temporal 180M~\cite{zellers2021merlot}, HowTo100M~\cite{miech2019howto100m}, CC3M~\cite{sharma2018conceptual}, CC12M~\cite{changpinyo2021cc12m}, COCO~\cite{lin2014microsoft}, VisualGenome~\cite{krishna2017visual}, and SBU~\cite{ordonez2011im2text}.
When conducting text-to-video retrieval task, video-text matching loss and masked language modeling loss are adopted and the former one is used as the primary loss.

\subsection{Evaluation metrics}
For the video retrieval task, we report the standard retrieval metrics, Recall at K (R@K) metric (K=1,5,10) and Median Rank (MedR).
Accuracy metric is reported for video question answering task which includes both multi-choice and open-ended questions.
As for video captioning, BLEU~\cite{papineni2002bleu}, METEOR~\cite{banerjee2005meteor}, ROUGE-L~\cite{lin2004rouge}, and CIDEr~\cite{vedantam2015cider} are reported.

\subsection{MELTR Details.}
We use the Adam~\cite{kingma2015adam} optimizer with an initial learning rate $\alpha$ = 3e-5 and $\beta$ = 1e-4 with a linear learning rate decay strategy.
For MELTR, we use one transformer encoder layer with 8 attention heads and 512 hidden dimensions.
We trained 40, 20, and 20 epochs on the text-to-video retrieval, video question answering, and video captioning tasks with 8 $\times$ Tesla A100 GPUs, respectively.
We search $\gamma$ in $\{0.1, 0.3, 0.5\}$ for the regularization term and use $K = 3$ in Eq. (13) of the main paper.
\section{Dataset Details}

\noindent \textbf{YouCook2}. 
YouCook2~\cite{zhou2018towards} consists of 2k videos, which cover 89 types of recipes. 
Each video contains multiple video clips accompanied by text descriptions. 
The train dataset contains 1,261 samples, and the test set contains 439 samples, respectively.

\noindent \textbf{MSRVTT.}
The original MSRVTT-full~\cite{xu2016msr} dataset, used on video captioning task, contains 6,513 train, 497 validation, and 2,990 test samples.
However, we have observed a wide range of dataset split variations throughout research on text-to-video retrieval.
One split variant randomly samples 1,000 clip-text pairs from the test set for evaluation and uses the rest of the 9,000 samples as train data~\cite{yu2018joint}, which is commonly denoted as the 1kA split.
On the other hand, the 1kB split uses the identical 1,000 test split of 1kA for the test, whereas the train set is a subset of 1kA's containing 6,656 samples~\cite{miech2018learning}.
Another commonly used data split also uses the identical 1,000 test set, while adopting both the train and validation set from the standard MSRVTT for training.
We evaluated our method on two split protocols most prominently observed in the literature, 1kA, and 7k.
For convenience, we denote the former as MSRVTT-9k and the latter as MSRVTT-7k.

\noindent \textbf{TGIF-QA.}
TGIF-QA~\cite{jang2017tgif} contains 165k QA pairs of animated GIFs.
The dataset provides three different subtasks:  TGIF-Action, TGIF-Transition and TGIF-Frame.
TGIF-Action is to identify repeated actions, TGIF-Transition is to identify the transition between states, and TGIF-Frame is to answer questions given a GIF frame.
TGIF-Action and TGIF-Transition are conducted under the multi-choice question answering setting, predicting the best answer given five options.
TGIF-Frame is experimented as the open-ended question answering with 1,540 most frequent answer candidates.

\noindent \textbf{MSVD-QA.}
MSVD-QA~\cite{xu2017video} contains 47k open-ended questions on 2k videos, derived from the original MSVD dataset~\cite{chen2011collecting}.
We construct the answer set with 1,000 most frequently appeared answers.

\noindent \textbf{CMU-MOSI.}
For the multi-modal sentiment analysis task, we adopt the CMU-MOSI dataset~\cite{zadeh2016mosi} which consists of 2,199 opinion video clips annotated with sentiment intensity values from -3 to 3.

\begin{figure*}[t!] 
    \centering
    \includegraphics[width=0.98\textwidth]{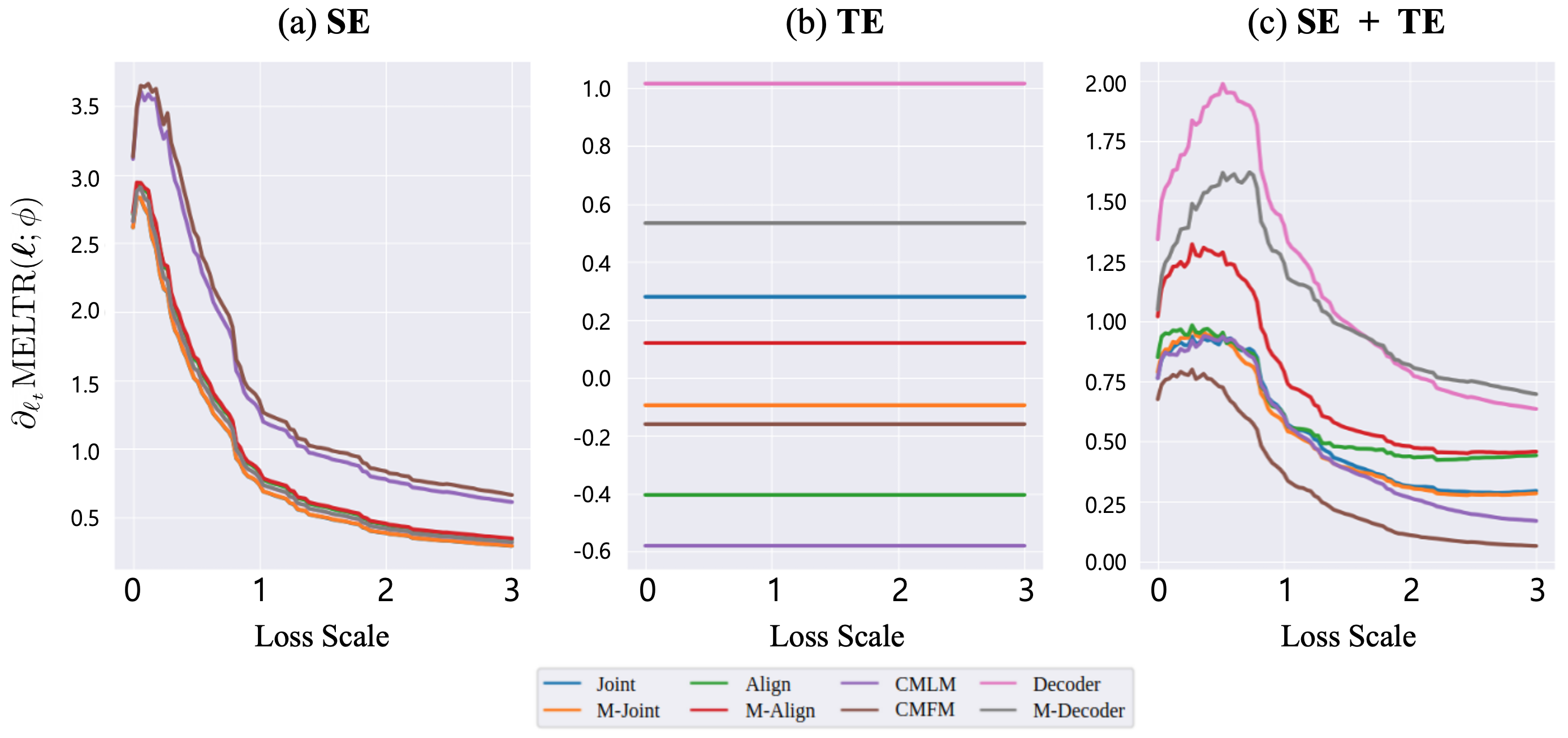}
    \caption{\textbf{Gradient by embedding type.} The gradient of MELTR output with respect to each task loss is plotted for different input embedding types. 
    (a) Gradient values are generally similar across tasks, and only those with distinct loss scales are distinguished. 
    (b) Gradients are different across tasks, but stay constant along loss scale, as loss scale information is not provided. 
    (c) MELTR learned to effectively consider both loss scale and task information.
    }
    \label{fig:apdx_grad}
\end{figure*}
\section{Effectiveness of the Regularization Term}

We proposed the regularization term $\Lreg$ in Section 4.1 of the main paper.
Eq. (6) of the main paper encourages the learned loss $\text{MELTR}(\Ell;\phi)$ to stay within a reasonable range to avoid meta-overfitting.
Table~\ref{tab:apdx_gamma} shows the ablation study for $\Lreg$ by adjusting the regularization strength $\gamma$ on the text-to-video retrieval of MSRVTT-7k.
Without the regularization term, \textit{i.e.}, $\gamma = 0$, it shows the performance of 27.6\% on R@1 metric.
The performance improves at $\gamma = 1$ or $\gamma = 10$ by a margin of 1\% than without $\Lreg$.
\section{Effectiveness of transformer architecture}

In this section, we conduct an ablation study for the architecture type of MELTR on the text-to-video retrieval on MSRVTT by replacing the transformer with a linear layer.
Table~\ref{tab:apdx_arch} demonstrates that the transformer architecture improves by margin of 1\% than the linear layer by taking advantage of the self-attention layer.
Furthermore, we use both the scale embedding and task embedding (\textbf{SE} + \textbf{TE}) as the input of MELTR.
Only with \textbf{SE}, MELTR cannot consider task information and hence the performance decreases.
However, only with \textbf{TE}, MELTR cannot be trained since the input losses are not passed to MELTR, \textit{i.e.}, $\nabla_w \aux$ is always zero.
\begin{figure*}[t!] 
    \centering
    \includegraphics[width=0.98\textwidth]{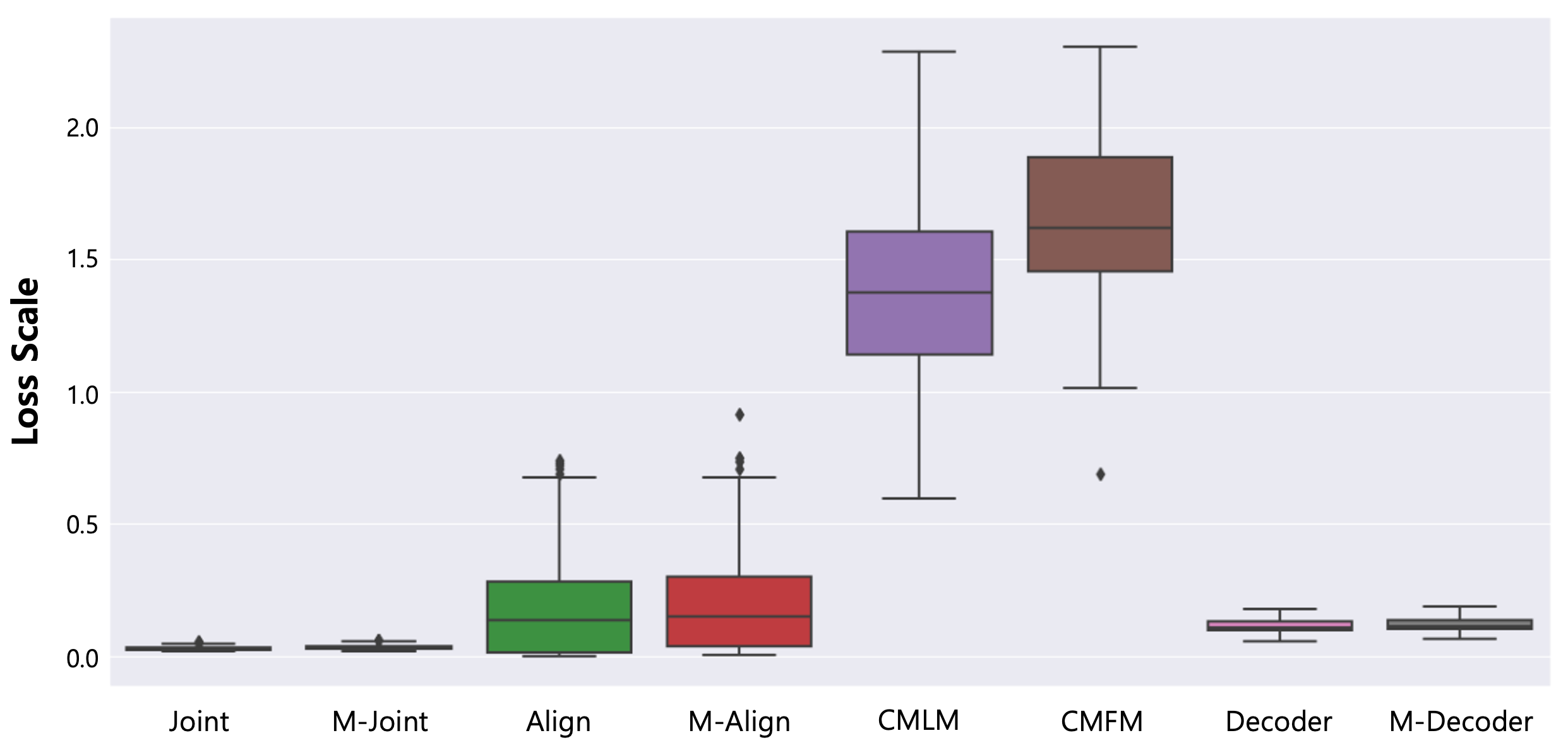}
    \caption{\textbf{Loss range for each task.} 
    The ranges of each task loss for each data sample are plotted. 
    A clear distinction is observed between the range of CMLM / CMFM loss and the rest of the task losses.}
    \label{fig:box}
\end{figure*}
\begin{table}[t!]
    \centering
    \setlength{\tabcolsep}{3.5pt}
    \caption{\textbf{Regularization strength.}}
    \begin{tabular}{c | c c c c c c c}
        \toprule
        $\boldsymbol{\gamma}$ & 0 & 0.001 & 0.01 & 0.1 & 1.0 & 10 & 100 \\
        \midrule
        R@1 & 27.6 & 27.8 & 28.1 & 28.4 & \textbf{28.6} & \textbf{28.6} & 28.5 \\
        \bottomrule
    \end{tabular}
    \label{tab:apdx_gamma}
\end{table}
\begin{table}[t!]
    \centering
    \setlength{\tabcolsep}{3.5pt}
    \caption{\textbf{The effect of MELTR architecture.} 
    Experimental results for different MELTR architectures are provided. 
    The performances are reported for video retrieval on MSRVTT. 
    We do not report performance for task-embedding-only Transformer, as our optimization method is not trained properly in such a setting; $\nabla_w \Laux$ is always zero.}
    \begin{tabular}{c c}
        \toprule
        $\textbf{Architecture}$ & \textbf{R@1} \\
        \midrule
        \midrule
        Linear & 27.6 \\
        \midrule
        Transformer (SE+TE) & \textbf{28.6} \\
        Transformer (SE only) & 27.9 \\
        Transformer (TE only) & - \\
        \bottomrule
    \end{tabular}
    \label{tab:apdx_arch}
\end{table}
\section{Effectiveness of input type}
In this section, we provide a qualitative analysis for each input type (\textbf{SE} only, \textbf{TE} only, and \textbf{SE} + \textbf{TE}).
We visualize $\partial_{\ell_t} \text{MELTR}(\Ell;\phi)$ denoted in Section 5.2 of the main paper.
We calculate it in the same way as in the main paper for three input types on the video captioning task of YouCook2.

Figure~\ref{fig:apdx_grad} illustrates $\partial_{\ell_t} \text{MELTR}(\Ell;\phi)$ with respect to the scales of the input loss values.
When only the \textbf{SE} is fed in Figure~\ref{fig:apdx_grad}(a), MELTR tends to focus on reasonably challenging samples and downweight the noisy samples as discussed in Section 5.2 of the main paper.
Also note that without task information, we observe that the tendency is separated into two clusters with respect to $\partial_{\ell_t} \text{MELTR}(\Ell;\phi)$: ($\Lcmlm$, $\Lcmfm$) and ($\Ljoint$, $\Lmjoint$, $\Lalign$, $\Lmalign$, $\Ldecoder$, $\Lmdecoder$).
We believe that this is because the auxiliary losses are grouped based on the ranges of each loss, as seen in Figure~\ref{fig:box}, and MELTR distinguishes the tasks to some extent by learning the range of losses without the \textbf{TE}.
As for the \textbf{TE} in Figure~\ref{fig:apdx_grad}(b), $\partial_{\ell_t} \text{MELTR}(\Ell;\phi)$ is obviously invariant to the scale of losses and depend only on the task types.
$\Ldecoder$ and $\Lmdecoder$ rank high because they improve the performance on the video captioning task.
In Figure~\ref{fig:apdx_grad}(c), MELTR finally takes into account the tasks which are advantageous on the primary task, and guides a learner to focus on a reasonably challenging samples as discussed in Section 5.2 of the main paper, when using the summation of two embeddings (\textbf{SE} + \textbf{TE}).
\begin{figure}[t!] 
    \centering
    \includegraphics[width=0.49\textwidth]{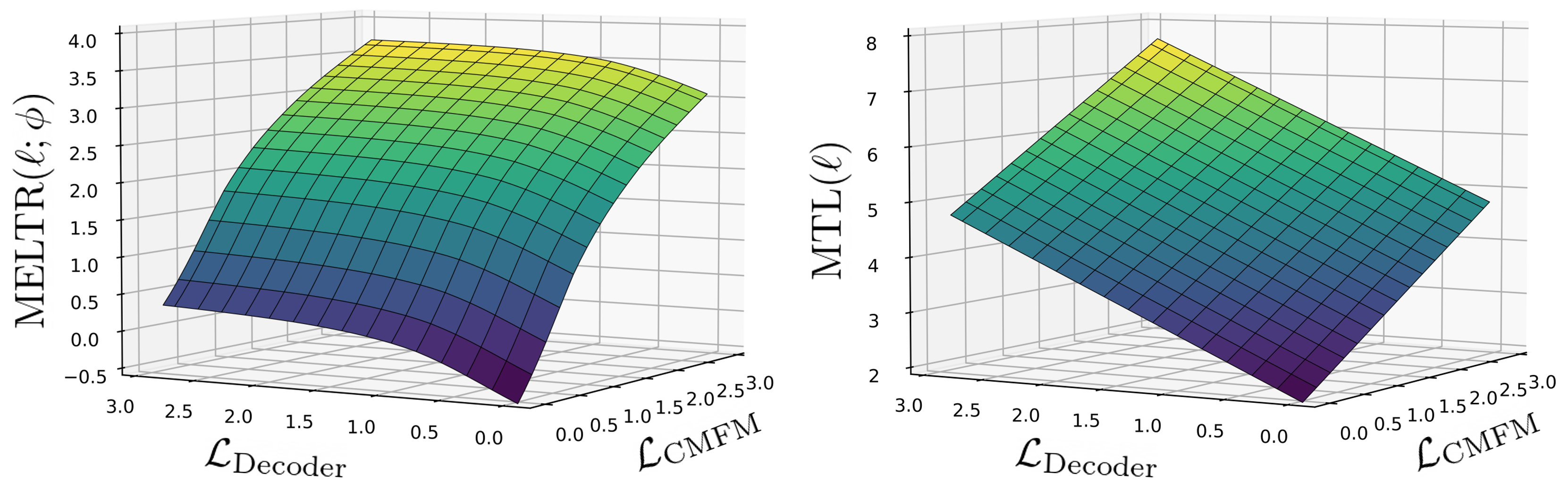}
    \caption{\textbf{Non-linearity of MELTR.} 
    MELTR (left) and MTL (right) output with respect to $\Ldecoder$ and $\Lcmfm$.
    }
    \label{fig:nonlinearity}
\end{figure}
\section{Non-linearity of MELTR}

MELTR provides more flexible and effective transformations beyond a simple linear combination of losses through transformer architecture.
Table 8 of the main paper evidences that MELTR outperforms two linear combinations, the sum of losses (multi-task learning, MTL) and an adaptive and learned linear combination (Meta-Weight Net), by 2.4 and 1.3 R@1 in MSRVTT for text-to-video retrieval.
Qualitatively, Figure~\ref{fig:nonlinearity} shows the non-linearity of MELTR in contrast to the multi-task learning (MTL) by visualizing their outputs given two input losses: $\Ldecoder$ and $\Lcmfm$.
\begin{table}[t!]
    \centering
    \small
    \caption{\footnotesize \textbf{Video captioning on YouCook2.}
    B3, B4, M, and R mean BLEU-3, BLEU-4, METEOR, and ROUGE-L, respectively.
    `Ori.' contains original five auxiliary losses: $\mathcal{L}_\text{Joint}$, $\mathcal{L}_\text{Align}$, $\mathcal{L}_\text{CMLM}$, $\mathcal{L}_\text{CMFM}$, and $\mathcal{L}_\text{Decoder}$. 
    Also, the last column reports the averaged gain across metrics compared to the Ori. settings of {\color{blue}MTL} and {\color{red}MELTR}, respectively.
    }
    \begin{adjustbox}{width=0.49\textwidth}
    \renewcommand{\arraystretch}{1.3}
    \begin{tabular}{c|c|c c c c | c}
        \specialrule{.2em}{.1em}{.1em} 
        \textbf{Auxiliary losses} & \textbf{Training} & \textbf{B3} & \textbf{B4} & \textbf{M} & \textbf{R} & \textbf{avg. gain} \\
        \specialrule{.2em}{.1em}{.1em} 
        \multirow{2}{*}{Ori.} & MTL & 20.68 & 14.95 & 20.18 & 44.25 & \color{blue}+0.00 \\
        \cline{2-7}
        & METLR & \textbf{23.47} & \textbf{17.29} & \textbf{22.25} & \textbf{45.67} & \color{red}+0.00 \\
        \specialrule{.2em}{.1em}{.1em} 
        \multirow{2}{*}{Ori. + $\mathcal{L}_\text{M-Decoder}$} & MTL & 21.51 & 15.69 & 20.73 & 45.05 & \color{blue}+0.73 \\
        \cline{2-7}
        & MELTR & \textbf{23.86} & \textbf{17.59} & \textbf{22.34} & \textbf{46.76} & \color{red}+0.47 \\
        \specialrule{.2em}{.1em}{.1em} 
        \multirow{2}{*}{Ori. + $\mathcal{L}_\text{M-Joint}$} & MTL & 21.00 & 15.19 & 20.46 & 44.63 & \color{blue}+0.31 \\
        \cline{2-7}
        & MELTR & \textbf{23.76} & \textbf{17.53} & \textbf{22.22} & \textbf{46.63} & \color{red}+0.37 \\
        \specialrule{.2em}{.1em}{.1em}
        \multirow{2}{*}{Ori. + $\mathcal{L}_\text{M-Align}$} & MTL & 20.76 & 15.01 & 20.27 & 44.29 & \color{blue}+0.07 \\
        \cline{2-7}
        & MELTR & \textbf{23.55} & \textbf{17.45} & \textbf{22.16} & \textbf{46.56} & \color{red}+0.26 \\
        \specialrule{.2em}{.1em}{.1em}
        \multirow{2}{*}{\parbox{2.3cm}{Ori. + $\mathcal{L}_\text{M-Decoder}$ + $\mathcal{L}_\text{M-Align}$ + $\mathcal{L}_\text{M-Joint}$}} & MTL & 21.72 & 15.93 & 20.89 & 45.16 & \color{blue}+0.91 \\
        \cline{2-7}
        & MELTR & \textbf{24.12} & \textbf{17.92} & \textbf{22.56} & \textbf{47.04} & \color{red}+0.74 \\
        \specialrule{.2em}{.1em}{.1em}
    \end{tabular}
    \end{adjustbox}
    \label{tab:univl_advanced}
\end{table}
\begin{table*}[t!]
    \centering
    \setlength{\tabcolsep}{3.5pt}
    \caption{\textbf{Additional quantitative results.}
    (Left) The accuracy of video question answering on MSVD-QA is reported.
    (Middle) The accuracy of action recognition on Kinetics400 is reported.
    (Right) The accuracy of image classification on CIFAR-100 is reported.
    }
    \begin{tabular}[t]{l|c}
        \toprule
        Models & Accuracy \\
        \midrule
        \midrule
        ALPRO & 45.9 \\
        ALPRO + \textbf{MELTR} & \textbf{46.8} \\
        \bottomrule
    \end{tabular}
    \hspace{1cm}
    \begin{tabular}[t]{l|c}
        \toprule
        Models & Accuracy \\
        \midrule
        \midrule
        Violet & 72.4 \\
        Violet + \textbf{MELTR} & \textbf{73.1} \\
        \bottomrule
    \end{tabular}
    \hspace{1cm}
    \begin{tabular}[t]{l|c}
        \toprule
        Models & Accuracy \\
        \midrule
        \midrule
        ResNet32 & 66.5 \\
        ResNet32 + \textbf{MELTR} & \textbf{69.2} \\
        \bottomrule
    \end{tabular}
    \label{tab:apdx_quan}
\end{table*}
\section{Effectiveness of advanced loss of UniVL}

For video captioning on YouCook2, in order of importance, the losses can be sorted as $\mathcal{L}_\text{M-Decoder}$, $\mathcal{L}_\text{M-Joint}$, and $\mathcal{L}_\text{M-Align}$. 
Table~\ref{tab:univl_advanced} shows the additional ablation study on newly added losses.
First, using all three newly added losses improves the performance with both MTL (+0.91) and MELTR (+0.74) on average.
As for the individual loss, by adding $\mathcal{L}_\text{M-Decoder}$, the average performance gain of MELTR is 0.47.
On the other hand, with $\mathcal{L}_\text{M-Joint}$ or $\mathcal{L}_\text{M-Align}$, the performance gap is decreased to 0.37 and 0.26 respectively, implying that they are relatively less effective for video captioning than $\mathcal{L}_\text{M-Decoder}$ as observed in Sec. 5.2 of the main paper. 
\section{Adaptation to a new baseline and tasks}

\noindent \textbf{Plug-in to a new baseline.}
In Table~\ref{tab:apdx_quan} (Left), we conduct an experiment with another strong model ALPRO~\cite{li2022align} trained with four pretext losses.
In the video question answering task on MSVD-QA, ALPRO shows the original performance of 45.9\%, and MELTR improves it to 46.8\%. 

\noindent \textbf{Video only setting.} 
We also evaluate action recognition performance on Kinetics400~\cite{carreira2017quo} by applying MELTR to Violet in Table~\ref{tab:apdx_quan} (Middle).
Since the action recognition is a unimodal task with \emph{only} `videos', we use the following two losses: classification loss (primary task) and Masked Visual-token Modeling loss (MVM; auxiliary task).
Violet's accuracy is improved from 72.4\% to 73.1\%.

\noindent \textbf{Image only setting.} 
Furthermore, to verify the generalizability of MELTR to other domains, we also conduct our experiment on the  `image' domain (image classification on CIFAR-100) with ResNet32 backbone in Table~\ref{tab:apdx_quan} (Right).
We add two simple auxiliary losses (mixup~\cite{zhang2017mixup} and rotation~\cite{chen2020simple}) with a basic classification loss.
Our MELTR outperforms the baseline by a margin of 2.7\%.
These experimental results demonstrate that MELTR is a general framework to be adapted to a wide range of domains and tasks.

%%%%%%%%% REFERENCES
{\small
\bibliographystyle{unsrt}
\bibliography{egbib}

\begin{thebibliography}{10}

\bibitem{bommasani2021opportunities}
Rishi Bommasani, Drew~A Hudson, Ehsan Adeli, Russ Altman, Simran Arora, Sydney
  von Arx, Michael~S Bernstein, Jeannette Bohg, Antoine Bosselut, Emma
  Brunskill, et~al.
\newblock On the opportunities and risks of foundation models, 2021.

\bibitem{bertasius2021space}
Gedas Bertasius, Heng Wang, and Lorenzo Torresani.
\newblock Is space-time attention all you need for video understanding?
\newblock In {\em ICML}, 2021.

\bibitem{brown2020language}
Tom Brown, Benjamin Mann, Nick Ryder, Melanie Subbiah, Jared~D Kaplan, Prafulla
  Dhariwal, Arvind Neelakantan, Pranav Shyam, Girish Sastry, Amanda Askell,
  Sandhini Agarwal, Ariel Herbert-Voss, Gretchen Krueger, Tom Henighan, Rewon
  Child, Aditya Ramesh, Daniel Ziegler, Jeffrey Wu, Clemens Winter, Chris
  Hesse, Mark Chen, Mateusz Sigler, Scott Gray, Benjamin Chess, Jack Clark,
  Christopher Berner, Sam McCandlish, Alec Radford, Ilya Sutskever, and Dario
  Amodei.
\newblock Language models are few-shot learners.
\newblock In {\em NeurIPS}, 2020.

\bibitem{radford2021learning}
Alec Radford, Jong~Wook Kim, Chris Hallacy, Aditya Ramesh, Gabriel Goh,
  Sandhini Agarwal, Girish Sastry, Amanda Askell, Pamela Mishkin, Jack Clark,
  Gretchen Krueger, and Ilya Sutskever.
\newblock Learning transferable visual models from natural language
  supervision.
\newblock In {\em ICML}, 2021.

\bibitem{jia2021scaling}
Chao Jia, Yinfei Yang, Ye~Xia, Yi-Ting Chen, Zarana Parekh, Hieu Pham, Quoc Le,
  Yun-Hsuan Sung, Zhen Li, and Tom Duerig.
\newblock Scaling up visual and vision-language representation learning with
  noisy text supervision.
\newblock In {\em ICML}, 2021.

\bibitem{akbari2021vatt}
Hassan Akbari, Liangzhe Yuan, Rui Qian, Wei-Hong Chuang, Shih-Fu Chang, Yin
  Cui, and Boqing Gong.
\newblock Vatt: Transformers for multimodal self-supervised learning from raw
  video, audio and text.
\newblock In {\em NeurIPS}, 2021.

\bibitem{luo2020univl}
Huaishao Luo, Lei Ji, Botian Shi, Haoyang Huang, Nan Duan, Tianrui Li, Jason
  Li, Taroon Bharti, and Ming Zhou.
\newblock Univl: A unified video and language pre-training model for multimodal
  understanding and generation, 2020.

\bibitem{zellers2021merlot}
Rowan Zellers, Ximing Lu, Jack Hessel, Youngjae Yu, Jae~Sung Park, Jize Cao,
  Ali Farhadi, and Yejin Choi.
\newblock Merlot: Multimodal neural script knowledge models.
\newblock In {\em NeurIPS}, 2021.

\bibitem{liebel2018auxiliary}
Lukas Liebel and Marco K{\"o}rner.
\newblock Auxiliary tasks in multi-task learning.
\newblock {\em arXiv preprint arXiv:1805.06334}, 2018.

\bibitem{Toshniwal2017}
Shubham Toshniwal, Hao Tang, Liang Lu, and Karen Livescu.
\newblock Multitask learning with low-level auxiliary tasks for encoder-decoder
  based speech recognition.
\newblock In {\em Interspeech}, 2017.

\bibitem{liu2019self}
Shikun Liu, Andrew Davison, and Edward Johns.
\newblock Self-supervised generalisation with meta auxiliary learning.
\newblock In {\em NeurIPS}, 2019.

\bibitem{navon2020auxiliary}
Aviv Navon, Idan Achituve, Haggai Maron, Gal Chechik, and Ethan Fetaya.
\newblock Auxiliary learning by implicit differentiation.
\newblock In {\em ICLR}, 2021.

\bibitem{shu2019meta}
Jun Shu, Qi~Xie, Lixuan Yi, Qian Zhao, Sanping Zhou, Zongben Xu, and Deyu Meng.
\newblock Meta-weight-net: Learning an explicit mapping for sample weighting.
\newblock In {\em NeurIPS}, 2019.

\bibitem{vaswani2017attention}
Ashish Vaswani, Noam Shazeer, Niki Parmar, Jakob Uszkoreit, Llion Jones,
  Aidan~N Gomez, Lukasz Kaiser, and Illia Polosukhin.
\newblock Attention is all you need.
\newblock In {\em NeurIPS}, 2017.

\bibitem{hwang2020self}
Dasol Hwang, Jinyoung Park, Sunyoung Kwon, KyungMin Kim, Jung-Woo Ha, and
  Hyunwoo~J Kim.
\newblock Self-supervised auxiliary learning with meta-paths for heterogeneous
  graphs.
\newblock In {\em NeurIPS}, 2020.

\bibitem{fu2021violet}
Tsu-Jui Fu, Linjie Li, Zhe Gan, Kevin Lin, William~Yang Wang, Lijuan Wang, and
  Zicheng Liu.
\newblock Violet: End-to-end video-language transformers with masked
  visual-token modeling.
\newblock {\em arXiv preprint arXiv:2111.12681}, 2021.

\bibitem{wang2022all}
Alex~Jinpeng Wang, Yixiao Ge, Rui Yan, Yuying Ge, Xudong Lin, Guanyu Cai,
  Jianping Wu, Ying Shan, Xiaohu Qie, and Mike~Zheng Shou.
\newblock All in one: Exploring unified video-language pre-training.
\newblock {\em arXiv preprint arXiv:2203.07303}, 2022.

\bibitem{chen2021evaluating}
Mark Chen, Jerry Tworek, Heewoo Jun, Qiming Yuan, Henrique~Ponde
  de~Oliveira~Pinto, Jared Kaplan, Harri Edwards, Yuri Burda, Nicholas Joseph,
  Greg Brockman, Alex Ray, Raul Puri, Gretchen Krueger, Michael Petrov, Heidy
  Khlaaf, Girish Sastry, Pamela Mishkin, Brooke Chan, Scott Gray, Nick Ryder,
  Mikhail Pavlov, Alethea Power, Lukasz Kaiser, Mohammad Bavarian, Clemens
  Winter, Philippe Tillet, Felipe~Petroski Such, Dave Cummings, Matthias
  Plappert, Fotios Chantzis, Elizabeth Barnes, Ariel Herbert-Voss,
  William~Hebgen Guss, Alex Nichol, Alex Paino, Nikolas Tezak, Jie Tang, Igor
  Babuschkin, Suchir Balaji, Shantanu Jain, William Saunders, Christopher
  Hesse, Andrew~N. Carr, Jan Leike, Josh Achiam, Vedant Misra, Evan Morikawa,
  Alec Radford, Matthew Knight, Miles Brundage, Mira Murati, Katie Mayer, Peter
  Welinder, Bob McGrew, Dario Amodei, Sam McCandlish, Ilya Sutskever, and
  Wojciech Zaremba.
\newblock Evaluating large language models trained on code, 2021.

\bibitem{devlin2019bert}
Jacob Devlin, Ming-Wei Chang, Kenton Lee, and Kristina Toutanova.
\newblock Bert: Pre-training of deep bidirectional transformers for language
  understanding.
\newblock In {\em NACCL}, 2019.

\bibitem{gabeur2020multi}
Valentin Gabeur, Chen Sun, Karteek Alahari, and Cordelia Schmid.
\newblock Multi-modal transformer for video retrieval.
\newblock In {\em ECCV}, 2020.

\bibitem{neimark2021video}
Daniel Neimark, Omri Bar, Maya Zohar, and Dotan Asselmann.
\newblock Video transformer network.
\newblock In {\em ICCVW}, 2021.

\bibitem{tan2021look}
Reuben Tan, Bryan~A. Plummer, Kate Saenko, Hailin Jin, and Bryan Russell.
\newblock Look at what i’m doing: Self-supervised spatial grounding of
  narrations in instructional videos.
\newblock In {\em NeurIPS}, 2021.

\bibitem{wei2020multi}
Xi~Wei, Tianzhu Zhang, Yan Li, Yongdong Zhang, and Feng Wu.
\newblock Multi-modality cross attention network for image and sentence
  matching.
\newblock In {\em CVPR}, 2020.

\bibitem{yang2021taco}
Jianwei Yang, Yonatan Bisk, and Jianfeng Gao.
\newblock Taco: Token-aware cascade contrastive learning for video-text
  alignment.
\newblock In {\em ICCV}, 2021.

\bibitem{kuang2021video}
Haofei Kuang, Yi~Zhu, Zhi Zhang, Xinyu Li, Joseph Tighe, S\"oren Schwertfeger,
  Cyrill Stachniss, and Mu~Li.
\newblock Video contrastive learning with global context.
\newblock In {\em ICCVW}, 2021.

\bibitem{li2020hero}
Linjie Li, Yen-Chun Chen, Yu~Cheng, Zhe Gan, Licheng Yu, and Jingjing Liu.
\newblock Hero: Hierarchical encoder for video+language omni-representation
  pre-training, 2020.

\bibitem{sun2019learning}
Chen Sun, Fabien Baradel, Kevin Murphy, and Cordelia Schmid.
\newblock Learning video representations using contrastive bidirectional
  transformer, 2019.

\bibitem{sun2019videobert}
Chen Sun, Austin Myers, Carl Vondrick, Kevin Murphy, and Cordelia Schmid.
\newblock Videobert: A joint model for video and language representation
  learning.
\newblock {\em CoRR}, 2019.

\bibitem{zhu2020actbert}
Linchao Zhu and Yi~Yang.
\newblock Actbert: Learning global-local video-text representations.
\newblock In {\em CVPR}, 2020.

\bibitem{ruder2017overview}
Sebastian Ruder.
\newblock An overview of multi-task learning in deep neural networks.
\newblock {\em arXiv preprint arXiv:1706.05098}, 2017.

\bibitem{xu2021leveraging}
Lian Xu, Wanli Ouyang, Mohammed Bennamoun, Farid Boussaid, Ferdous Sohel, and
  Dan Xu.
\newblock Leveraging auxiliary tasks with affinity learning for weakly
  supervised semantic segmentation.
\newblock In {\em ICCV}, 2021.

\bibitem{ye2021auxiliary}
Joel Ye, Dhruv Batra, Abhishek Das, and Erik Wijmans.
\newblock Auxiliary tasks and exploration enable objectnav, 2021.

\bibitem{baydin2018automatic}
Atilim~Gunes Baydin, Barak~A Pearlmutter, Alexey~Andreyevich Radul, and
  Jeffrey~Mark Siskind.
\newblock Automatic differentiation in machine learning: a survey.
\newblock In {\em JMLR}, 2018.

\bibitem{derivatives2000principles}
Evaluating Derivatives.
\newblock Principles and techniques of algorithmic differentiation.
\newblock {\em SIAM}, 2000.

\bibitem{miech2019howto100m}
Antoine Miech, Dimitri Zhukov, Jean-Baptiste Alayrac, Makarand Tapaswi, Ivan
  Laptev, and Josef Sivic.
\newblock Howto100m: Learning a text-video embedding by watching hundred
  million narrated video clips.
\newblock In {\em ICCV}, 2019.

\bibitem{grazzi2020iteration}
Riccardo Grazzi, Luca Franceschi, Massimiliano Pontil, and Saverio Salzo.
\newblock On the iteration complexity of hypergradient computation.
\newblock In {\em ICML}, 2020.

\bibitem{finn2017model}
Chelsea Finn, Pieter Abbeel, and Sergey Levine.
\newblock Model-agnostic meta-learning for fast adaptation of deep networks.
\newblock In {\em ICML}, 2017.

\bibitem{franceschi2017forward}
Luca Franceschi, Michele Donini, Paolo Frasconi, and Massimiliano Pontil.
\newblock Forward and reverse gradient-based hyperparameter optimization.
\newblock In {\em ICML}, 2017.

\bibitem{franceschi2018bilevel}
Luca Franceschi, Paolo Frasconi, Saverio Salzo, Riccardo Grazzi, and
  Massimiliano Pontil.
\newblock Bilevel programming for hyperparameter optimization and
  meta-learning.
\newblock In {\em ICML}, 2018.

\bibitem{maclaurin2015gradient}
Dougal Maclaurin, David Duvenaud, and Ryan Adams.
\newblock Gradient-based hyperparameter optimization through reversible
  learning.
\newblock In {\em ICML}, 2015.

\bibitem{lorraine2020optimizing}
Jonathan Lorraine, Paul Vicol, and David Duvenaud.
\newblock Optimizing millions of hyperparameters by implicit differentiation.
\newblock In {\em AISTATS}, 2020.

\bibitem{pedregosa2016hyperparameter}
Fabian Pedregosa.
\newblock Hyperparameter optimization with approximate gradient.
\newblock In {\em ICML}, 2016.

\bibitem{rajeswaran2019meta}
Aravind Rajeswaran, Chelsea Finn, Sham~M Kakade, and Sergey Levine.
\newblock Meta-learning with implicit gradients.
\newblock {\em NeurIPS}, 2019.

\bibitem{antoniou2018train}
Antreas Antoniou, Harrison Edwards, and Amos Storkey.
\newblock How to train your maml.
\newblock In {\em ICLR}, 2019.

\bibitem{zintgraf2019fast}
Luisa Zintgraf, Kyriacos Shiarli, Vitaly Kurin, Katja Hofmann, and Shimon
  Whiteson.
\newblock Fast context adaptation via meta-learning.
\newblock In {\em ICML}, 2019.

\bibitem{zhou2018towards}
Luowei Zhou, Chenliang Xu, and Jason~J Corso.
\newblock Towards automatic learning of procedures from web instructional
  videos.
\newblock In {\em AAAI}, 2018.

\bibitem{xu2016msr}
Jun Xu, Tao Mei, Ting Yao, and Yong Rui.
\newblock Msr-vtt: A large video description dataset for bridging video and
  language.
\newblock In {\em CVPR}, 2016.

\bibitem{jang2017tgif}
Yunseok Jang, Yale Song, Youngjae Yu, Youngjin Kim, and Gunhee Kim.
\newblock Tgif-qa: Toward spatio-temporal reasoning in visual question
  answering.
\newblock In {\em CVPR}, 2017.

\bibitem{xu2017video}
Dejing Xu, Zhou Zhao, Jun Xiao, Fei Wu, Hanwang Zhang, Xiangnan He, and Yueting
  Zhuang.
\newblock Video question answering via gradually refined attention over
  appearance and motion.
\newblock In {\em MM}, 2017.

\bibitem{zadeh2016mosi}
Amir Zadeh, Rowan Zellers, Eli Pincus, and Louis-Philippe Morency.
\newblock Mosi: multimodal corpus of sentiment intensity and subjectivity
  analysis in online opinion videos.
\newblock {\em arXiv preprint arXiv:1606.06259}, 2016.

\bibitem{klein2015associating}
Benjamin Klein, Guy Lev, Gil Sadeh, and Lior Wolf.
\newblock Associating neural word embeddings with deep image representations
  using fisher vectors.
\newblock In {\em CVPR}, 2015.

\bibitem{miech2020end}
Antoine Miech, Jean-Baptiste Alayrac, Lucas Smaira, Ivan Laptev, Josef Sivic,
  and Andrew Zisserman.
\newblock {E}nd-to-{E}nd {L}earning of {V}isual {R}epresentations from
  {U}ncurated {I}nstructional {V}ideos.
\newblock In {\em CVPR}, 2020.

\bibitem{ging2020coot}
Simon Ging, Mohammadreza Zolfaghari, Hamed Pirsiavash, and Thomas Brox.
\newblock Coot: Cooperative hierarchical transformer for video-text
  representation learning.
\newblock In {\em NeurIPS}, 2020.

\bibitem{xu2021videoclip}
Hu~Xu, Gargi Ghosh, Po-Yao Huang, Dmytro Okhonko, Armen Aghajanyan, Florian
  Metze, Luke Zettlemoyer, and Christoph Feichtenhofer.
\newblock Videoclip: Contrastive pre-training for zero-shot video-text
  understanding.
\newblock In {\em EMNLP}, 2021.

\bibitem{yu2018joint}
Youngjae Yu, Jongseok Kim, and Gunhee Kim.
\newblock A joint sequence fusion model for video question answering and
  retrieval.
\newblock In {\em ECCV}, 2018.

\bibitem{lei2021clipbert}
Jie Lei, Linjie Li, Luowei Zhou, Zhe Gan, Tamara~L. Berg, Mohit Bansal, and
  Jingjing Liu.
\newblock Less is more: Clipbert for video-and-language learning via sparse
  sampling.
\newblock In {\em CVPR}, 2021.

\bibitem{wang2021t2vlad}
Xiaohan Wang, Linchao Zhu, and Yi~Yang.
\newblock T2vlad: global-local sequence alignment for text-video retrieval.
\newblock In {\em CVPR}, 2021.

\bibitem{bain2021frozen}
Max Bain, Arsha Nagrani, G{\"u}l Varol, and Andrew Zisserman.
\newblock Frozen in time: A joint video and image encoder for end-to-end
  retrieval.
\newblock In {\em ICCV}, 2021.

\bibitem{fan2019heterogeneous}
Chenyou Fan, Xiaofan Zhang, Shu Zhang, Wensheng Wang, Chi Zhang, and Heng
  Huang.
\newblock Heterogeneous memory enhanced multimodal attention model for video
  question answering.
\newblock In {\em CVPR}, 2019.

\bibitem{le2020hierarchical}
Thao~Minh Le, Vuong Le, Svetha Venkatesh, and Truyen Tran.
\newblock Hierarchical conditional relation networks for video question
  answering.
\newblock In {\em CVPR}, 2020.

\bibitem{jiang2020divide}
Jianwen Jiang, Ziqiang Chen, Haojie Lin, Xibin Zhao, and Yue Gao.
\newblock Divide and conquer: Question-guided spatio-temporal contextual
  attention for video question answering.
\newblock In {\em AAAI}, 2020.

\bibitem{lei2021less}
Jie Lei, Linjie Li, Luowei Zhou, Zhe Gan, Tamara~L Berg, Mohit Bansal, and
  Jingjing Liu.
\newblock Less is more: Clipbert for video-and-language learning via sparse
  sampling.
\newblock In {\em CVPR}, 2021.

\bibitem{zhou2018end}
L.~Zhou, Y.~Zhou, J.~J. Corso, R.~Socher, and C.~Xiong.
\newblock End-to-end dense video captioning with masked transformer.
\newblock In {\em CVPR}, 2018.

\bibitem{shi2019dense}
Botian Shi, Lei Ji, Yaobo Liang, Nan Duan, Peng Chen, Zhendong Niu, and Ming
  Zhou.
\newblock Dense procedure captioning in narrated instructional videos.
\newblock In {\em ACL}, 2019.

\bibitem{hessel2019case}
Jack Hessel, Bo~Pang, Zhenhai Zhu, and Radu Soricut.
\newblock A case study on combining asr and visual features for generating
  instructional video captions.
\newblock {\em arXiv preprint arXiv:1910.02930}, 2019.

\bibitem{chen2018less}
Yangyu Chen, Shuhui Wang, Weigang Zhang, and Qingming Huang.
\newblock Less is more: Picking informative frames for video captioning.
\newblock In {\em ECCV}, 2018.

\bibitem{pei2019marn}
Wenjie Pei, Jiyuan Zhang, Xiangrong Wang, Lei Ke, Xiaoyong Shen, and Yu-Wing
  Tai.
\newblock Memory-attended recurrent network for video captioning.
\newblock {\em CVPR}, 2019.

\bibitem{liu2020sibnet}
Sheng Liu, Zhou Ren, and Junsong Yuan.
\newblock Sibnet: Sibling convolutional encoder for video captioning.
\newblock {\em TPAMI}, 2020.

\bibitem{zhang2019object}
Junchao Zhang and Yuxin Peng.
\newblock Object-aware aggregation with bidirectional temporal graph for video
  captioning.
\newblock In {\em CVPR}, 2019.

\bibitem{hou2019joint}
Jingyi Hou, Xinxiao Wu, Wentian Zhao, Jiebo Luo, and Yunde Jia.
\newblock Joint syntax representation learning and visual cue translation for
  video captioning.
\newblock In {\em ICCV}, 2019.

\bibitem{zhang2020object}
Ziqi Zhang, Yaya Shi, Chunfeng Yuan, Bing Li, Peijin Wang, Weiming Hu, and
  Zheng{-}Jun Zha.
\newblock Object relational graph with teacher-recommended learning for video
  captioning.
\newblock In {\em CVPR}, 2020.

\bibitem{ji2021bilevel}
Kaiyi Ji, Junjie Yang, and Yingbin Liang.
\newblock Bilevel optimization: Convergence analysis and enhanced design.
\newblock In {\em ICML}, 2021.

\bibitem{univl_github}
Huaishao Luo, Lei Ji, Botian Shi, Haoyang Huang, Nan Duan, Tianrui Li, Jason
  Li, Taroon Bharti, and Ming Zhou.
\newblock \url{https://github.com/microsoft/UniVL}, 2020.

\bibitem{violet_github}
Tsu-Jui Fu, Linjie Li, Zhe Gan, Kevin Lin, William~Yang Wang, Lijuan Wang, and
  Zicheng Liu.
\newblock \url{https://github.com/tsujuifu/pytorch_violet}, 2021.

\bibitem{sharma2018conceptual}
Piyush Sharma, Nan Ding, Sebastian Goodman, and Radu Soricut.
\newblock Conceptual captions: A cleaned, hypernymed, image alt-text dataset
  for automatic image captioning.
\newblock In {\em ACL}, 2018.

\bibitem{all_github}
Alex~Jinpeng Wang, Yixiao Ge, Rui Yan, Yuying Ge, Xudong Lin, Guanyu Cai,
  Jianping Wu, Ying Shan, Xiaohu Qie, and Mike~Zheng Shou.
\newblock \url{https://github.com/showlab/all-in-one}, 2022.

\bibitem{changpinyo2021cc12m}
Soravit Changpinyo, Piyush Sharma, Nan Ding, and Radu Soricut.
\newblock {Conceptual 12M}: Pushing web-scale image-text pre-training to
  recognize long-tail visual concepts.
\newblock In {\em CVPR}, 2021.

\bibitem{lin2014microsoft}
Tsung-Yi Lin, Michael Maire, Serge Belongie, James Hays, Pietro Perona, Deva
  Ramanan, Piotr Doll{\'a}r, and C~Lawrence Zitnick.
\newblock Microsoft coco: Common objects in context.
\newblock In {\em ECCV}, 2014.

\bibitem{krishna2017visual}
Ranjay Krishna, Yuke Zhu, Oliver Groth, Justin Johnson, Kenji Hata, Joshua
  Kravitz, Stephanie Chen, Yannis Kalantidis, Li-Jia Li, David~A Shamma, et~al.
\newblock Visual genome: Connecting language and vision using crowdsourced
  dense image annotations.
\newblock {\em IJCV}, 2017.

\bibitem{ordonez2011im2text}
Vicente Ordonez, Girish Kulkarni, and Tamara Berg.
\newblock Im2text: Describing images using 1 million captioned photographs.
\newblock In {\em NeurIPS}, 2011.

\bibitem{papineni2002bleu}
Kishore Papineni, Salim Roukos, Todd Ward, and Wei-Jing Zhu.
\newblock Bleu: a method for automatic evaluation of machine translation.
\newblock In {\em ACL}, 2002.

\bibitem{banerjee2005meteor}
Satanjeev Banerjee and Alon Lavie.
\newblock Meteor: An automatic metric for mt evaluation with improved
  correlation with human judgments.
\newblock In {\em ACLW}, 2005.

\bibitem{lin2004rouge}
Chin-Yew Lin.
\newblock Rouge: A package for automatic evaluation of summaries.
\newblock {\em Text summarization branches out}, 2004.

\bibitem{vedantam2015cider}
Ramakrishna Vedantam, C~Lawrence~Zitnick, and Devi Parikh.
\newblock Cider: Consensus-based image description evaluation.
\newblock In {\em CVPR}, 2015.

\bibitem{kingma2015adam}
Diederik~P. Kingma and Jimmy Ba.
\newblock Adam: {A} method for stochastic optimization.
\newblock In {\em ICLR}, 2015.

\bibitem{miech2018learning}
Antoine Miech, Ivan Laptev, and Josef Sivic.
\newblock Learning a text-video embedding from incomplete and heterogeneous
  data.
\newblock {\em arXiv preprint arXiv:1804.02516}, 2018.

\bibitem{chen2011collecting}
David Chen and William~B Dolan.
\newblock Collecting highly parallel data for paraphrase evaluation.
\newblock In {\em ACL}, 2011.

\bibitem{li2022align}
Dongxu Li, Junnan Li, Hongdong Li, Juan~Carlos Niebles, and Steven~CH Hoi.
\newblock Align and prompt: Video-and-language pre-training with entity
  prompts.
\newblock In {\em CVPR}, 2022.

\bibitem{carreira2017quo}
Joao Carreira and Andrew Zisserman.
\newblock Quo vadis, action recognition? a new model and the kinetics dataset.
\newblock In {\em CVPR}, 2017.

\bibitem{zhang2017mixup}
Hongyi Zhang, Moustapha Cisse, Yann~N Dauphin, and David Lopez-Paz.
\newblock mixup: Beyond empirical risk minimization.
\newblock In {\em ICLR}, 2018.

\bibitem{chen2020simple}
Ting Chen, Simon Kornblith, Mohammad Norouzi, and Geoffrey Hinton.
\newblock A simple framework for contrastive learning of visual
  representations.
\newblock In {\em ICML}, 2020.

\end{thebibliography}
}

\end{document}